\documentclass[journal,twoside]{IEEEtran}
\usepackage{cite}

\ifCLASSINFOpdf
   \usepackage[pdftex]{graphicx}
   \graphicspath{ {./figures/} }
   \DeclareGraphicsExtensions{.pdf,.jpg,.png}
\else
\fi
\usepackage{amsmath}
\usepackage{amsfonts}
\usepackage{amssymb}
\usepackage{tabularx}
\usepackage{hyperref}
\usepackage{stackengine}
\usepackage{booktabs,caption}
\usepackage[flushleft]{threeparttable}
\usepackage{subcaption}
\usepackage[ruled,vlined]{algorithm2e}

\usepackage{arydshln}
\usepackage{float}
\floatstyle{plaintop}
\restylefloat{table}

\usepackage{fancyhdr}
\usepackage{kantlipsum}
\usepackage{eso-pic}

\begin{document}

\AddToShipoutPictureBG*{%
	\AtPageLowerLeft{%
		\setlength\unitlength{1in}%
		\hspace*{\dimexpr0.5\paperwidth\relax}
		\makebox(0,0.63)[c]{1558-0016~\copyright2022 IEEE. This work has been accepted for publication in IEEE Transactions on Intelligent Transportation Systems.}
		\makebox(0,0.3)[c]{The published version can be accessed at \href{https://ieeexplore.ieee.org/document/9712213}{https://ieeexplore.ieee.org/document/9712213}. DOI: \href{https://doi.org/10.1109/TITS.2022.3149370}{10.1109/TITS.2022.3149370}}
}}

%
\title{Towards Compact Autonomous Driving Perception\\with Balanced Learning and Multi-sensor Fusion}
%
%
%


\author{
Oskar~Natan
and~Jun~Miura,~\IEEEmembership{Member,~IEEE}
\thanks{O. Natan is a doctoral student at the Department of Computer Science and Engineering, Toyohashi University of Technology, Aichi 441-8580, Japan. He is also a lecturer at the Department of Computer Science and Electronics, Gadjah Mada University, Yogyakarta 55281, Indonesia.\\{\tt\small oskar.natan.ao@tut.jp; oskarnatan@ugm.ac.id}}%
\thanks{J. Miura is a Professor at the Department of Computer Science and Engineering, Toyohashi University of Technology, Aichi 441-8580, Japan.\\{\tt\small jun.miura@tut.jp}}%
\thanks{Manuscript received April 13$^{th}$, 2021, revised September 25$^{th}$, 2021, December 11$^{th}$, 2021, and January 29$^{th}$, 2022, and accepted February 3$^{rd}$, 2022.}
}

%
%

\markboth{IEEE Transactions on Intelligent Transportation Systems,~Vol.~xx, No.~x, Mm~yyyy}{Natan and Miura: Towards Compact Autonomous Driving Perception with Balanced Learning and Multi-sensor Fusion}
%



\maketitle

\begin{abstract}
	We present a novel compact deep multi-task learning model to handle various autonomous driving perception tasks in one forward pass. The model performs multiple views of semantic segmentation, depth estimation, light detection and ranging (LiDAR) segmentation, and bird's eye view projection simultaneously without being supported by other models. We also provide an adaptive loss weighting algorithm to tackle the imbalanced learning issue that occurred due to plenty of given tasks. Through data pre-processing and intermediate sensor fusion techniques, the model can process and combine multiple input modalities retrieved from RGB cameras, dynamic vision sensors (DVS), and LiDAR placed at several positions on the ego vehicle. Therefore, a better understanding of a dynamically changing environment can be achieved. Based on the ablation study, the model variant trained with our proposed method achieves a better performance. Furthermore, a comparative study is also conducted to clarify its performance and effectiveness against the combination of some recent models. As a result, our model maintains better performance even with much fewer parameters. Hence, the model can inference faster with less GPU memory utilization. Moreover, the result tends to be consistent in 3 different CARLA simulation datasets and 1 real-world nuScenes-lidarseg dataset. To support future research, we share codes and other files publicly at \href{https://github.com/oskarnatan/compact-perception}{https://github.com/oskarnatan/compact-perception}.
\end{abstract}

\begin{IEEEkeywords}
Driving perception, Scene understanding, Sensor fusion, Multi-task learning, Adaptive loss weighting.
\end{IEEEkeywords}

%
\IEEEpeerreviewmaketitle

\section{Introduction} \label{intro}
%
%
%
%

 

Based on the functional perspective, processing step, and information flow, a complete autonomous driving vehicle system is composed of four main stages: perception, planning, control, and system supervision. The main objective of the perception stage is to understand the surrounding area of the ego vehicle by processing given data retrieved from the sensor. Once clear information is available, the system is ready to receive commands like goals or missions for the planning stage. Then, after the trajectory or navigation path is generated, any instruction related to the actuator can be executed in the control stage. Finally, system supervision is responsible to monitor all aspects of the vehicle and ensure that everything is working as planned\cite{AD_review0}\cite{AD_review1}. As the first stage in an autonomous driving system, the perception stage holds an important role in the environmental scene understanding, which is the foundation before making any further decisions\cite{ADPsurvey4}. However, there are plenty of problems or issues in the area of autonomous driving perception that are challenging the understanding capability of the system\cite{ADPsurvey1}. For instance, the environmental condition where the system is deployed can be varied such as the weather can be sunny, cloudy, foggy, rainy, or poor illumination at night. The situation on the road is also unpredictable as there are numerous vehicles and pedestrians along with their uncertain behavior on the street that is challenging the system's adaptability. Therefore, the system must be supported with multiple kinds of sensors to provide various information and cover each other's weakness\cite{SF_redundant}. For example, a system cannot rely on the RGB camera in poor illumination conditions as it may fail to capture enough information. Therefore, another sensor such as DVS, radar, and LiDAR can serve as alternatives\cite{alternate_cam0_lidar}\cite{alternate_cam1_radar}\cite{alternate_cam2_dvs1}. Thus, a proper technique to pre-process and combine multiple data modalities is also needed to meet the system needs. Moreover, a full scene understanding with multiple perspectives of views is also important to improve the system's capability. Hence, several sensors can be attached at several positions on the ego vehicle to capture multiple views of the surrounding\cite{ADPsurvey3}.

\begin{figure*}
	\centering
	\includegraphics[width=\textwidth]{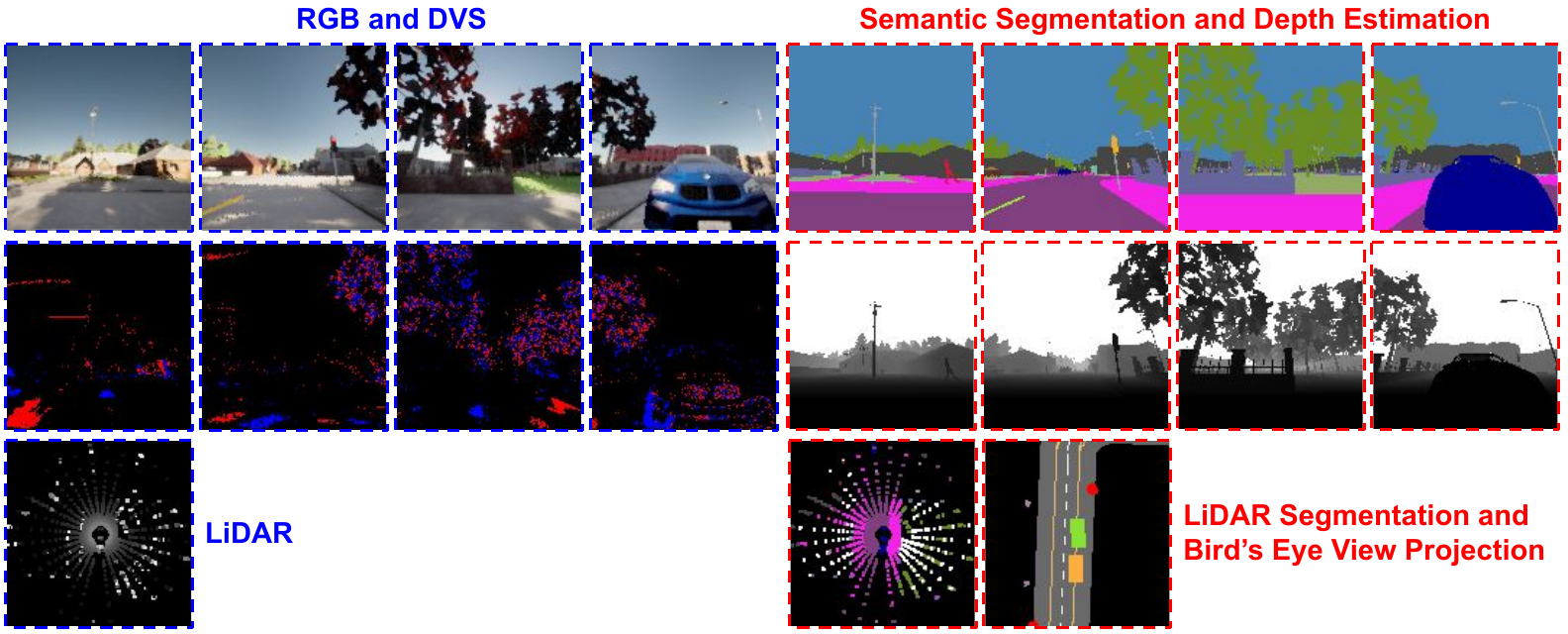}
	\caption{Overview of this research. Given four views (L-F-Ri-R) of RGB and DVS images, and a top view of pre-processed LiDAR point clouds as inputs (blue), the model performs four views of semantic segmentation and depth estimation along with a top view of LiDAR segmentation and bird's eye view projection as outputs (red) simultaneously in one forward pass.}
	\label{fig:carla_sample}
\end{figure*}

To achieve a compact scene understanding and fulfill the needs in the perception stage, we conduct research as shown in Figure \ref{fig:carla_sample}. Given a set of input data, we propose a model that performs various perception tasks with multiple perspectives of views: front (F), left (L), right (Ri), rear (R), and top (T). To be more specific, our model performs semantic segmentation (SS), depth estimation (DE), LiDAR segmentation (LS), and bird's eye view projection (BEVP) simultaneously. The model is supported with 4 RGB cameras, 4 DVS, and 1 LiDAR to provide rich information of a dynamically changing environment. Then, data pre-processing and sensor fusion techniques can be used to handle multiple kinds of data modalities\cite{ADPsurvey2}\cite{SF_survey}. Thus, a compact environmental scene understanding, especially at the surrounding area of the ego vehicle can be obtained. We consider using the multi-task learning (MTL) approach since handling each task with a single-task model can be very costly and inefficient\cite{ADPsurvey5}\cite{MTL4}\cite{MTL5}. However, for an MTL model, learning by combining several tasks is not always consistently better than in single-task learning. Different combined tasks may be conflicting with the gradient signals during the training process. If this issue is ignored, the outcome of the MTL approach cannot be optimal and cause performance degradation, or the training process may focus on one specific task only\cite{adaptloss0}. Hence, a proper strategy to balance the gradient and prevent imbalanced learning is a must. One of the possible answers is giving a set of loss weights to compensate for the imbalance. However, tuning a combination of loss weights can be tedious and computationally expensive. Therefore, rather than giving a fixed set of values\cite{adapt_lw2}, loss weights need to be tuned automatically\cite{LW_survey}\cite{adapt_lw1}.

Based on the aforementioned challenges in the field of environmental scene understanding, our approaches in achieving a compact autonomous driving perception and the novelty of this research can be summarized as follows:
\begin{itemize}
\item We present a compact deep MTL model that performs various driving perception tasks without being supported by other models. Our model performs multiple views of depth estimation, semantic segmentation, LiDAR segmentation, and bird's eye view projection simultaneously in one forward pass. Through data pre-processing and multi-sensor fusion techniques, the model can process various data modalities retrieved from 4 RGB cameras, 4 DVS, and 1 LiDAR to deal with a diverse and dynamically changing environment.
\item We provide an adaptive loss weighting strategy to tackle the imbalanced learning issue due to plenty of given tasks. To be more specific, we adopt and modify an algorithm called Gradient Normalization (GradNorm)\cite{adaptloss:gradnorm} to balance the learning process and show that the model is achieving better performance in many aspects. We perform an ablation study to understand the behavior and influence of this algorithm.
\item We conduct a comparative study between our best model variant and the combination of some recent models composed of both single-task and multi-task models that perform the same task. Based on the experiment result, we show that our model maintains a better performance even with much fewer parameters and smaller size. Therefore, our model can inference faster with less GPU utilization.
\end{itemize}

To strengthen our findings, we conduct experiments on four different datasets composed of three simulation datasets gathered using CARLA simulator\cite{carla} and one real-world dataset nuScenes-lidarseg\cite{nuscene}, which is also used to illustrate the implementation of the proposed model in a real-world scenario. The remainder of this paper is organized as follows. In Section \ref{relatedwork}, we review and summarize related research that inspires our works. In Section \ref{method}, we describe our proposed methods, including the model architecture and adaptive loss weighting algorithm. In Section \ref{expsetup}, we explain the dataset used for experiments and how the experiment is conducted. Then, we provide several points of ablation and comparative study in Section \ref{results}. Finally, the conclusion is presented in Section \ref{conclusion} followed by any possible research in the future.

\section{Related Work} \label{relatedwork}
In this section, we review several related works that are inspiring this research. We consider adopting and modifying some approaches to address challenges and issues in developing an autonomous driving perception model. We also summarize how our methods are built based on these works.


\subsection{Multi-task and Multi-modal Deep Learning} \label{multiio_ref}
The idea of learning multiple tasks simultaneously is to leverage shared features during the training process. In the area of multi-task learning (MTL) for autonomous driving perception, Lv et al.\cite{MTL0} develop a model that takes a single RGB image to predict lane area and lane marking simultaneously. With a simple encoder-decoder style, the model is made with one RGB encoder then branched into two task-specified decoders. A similar approach has been done by Chen et al.\cite{MTL1} where an MTL model called driving scene perception network is used to perform real-time joint detection, depth estimation, and semantic segmentation simultaneously. Moreover, Nakamura et al.\cite{MTL2} also conduct similar research to develop an MTL model that performs instance segmentation and depth estimation in one forward pass. Meanwhile, a different approach is presented by Yan et al. \cite{MTL3} where the model takes LiDAR point clouds to perform real-time occlusion-free road segmentation, dense road height estimation, and road topology recognition simultaneously. However, all of these approaches rely on one kind of input modality only which may fail in an unexpected environmental condition. Due to the diversity of environmental conditions, an autonomous driving agent cannot rely on one kind of sensor only. For example, during a poor illumination condition, an RGB camera will likely fail to capture the surrounding information. To address this issue, another kind of sensor can be used as an alternative in obtaining the information. Therefore, a sensor fusion strategy may be needed to combine various data representations.

In the field of sensor fusion, Muresan and Nedevschi \cite{SF_lidcam} combine LiDAR and RGB cameras to create affinity measurement and positional descriptor functions for autonomous driving agent to perform multi-object tracking. Pre-trained models are used to process LiDAR point clouds and RGB images separately to obtain both LS and SS images. Then, a hand-crafted feature extractor and aggregator are used to perform final calculation of object tracking. Another application of sensor fusion is presented by Dawar and Kehtarnavaz\cite{SF_depiner} where a depth camera and an inertial sensor are used for action detection and recognition in a continuous action stream. To extract depth images and inertial signal features, two separate deep learning-based encoders are used to process each input. A convolutional neural network (CNN) is used to handle the depth image, while a combination of CNN and long-short term memory network is used to process inertial signals. Each encoder is performing detection and recognition, then a separate decision fusion model is used to make the final decision by leveraging extracted features from each encoder. However, this kind of late fusion strategy can lose potentially useful information as the extracted features are not shared among the encoders. To address this issue, Nie et al.\cite{SF_interfusion} develop a multi-modality fusion framework called Integrated Multimodality Fusion Deep Neural Network (IMF-DNN) based on the intermediate fusion strategy where the extracted features are fused at some points at the network architecture. Their model takes multiple input modalities composed of LiDAR point clouds and RGB images, then fuses the extracted features several times. As a result, the IMF-DNN achieves higher performance in performing object detection and end-to-end driving policy in a diverse environment.

In this research, we imitate the architecture style presented by Lv et al.\cite{MTL0} that simply branches the decoder for each task. Hence, we will have a task-specific decoder for each task on each view. Then, we adopt the intermediate fusion strategy proposed by Nie et al.\cite{SF_interfusion} to process and combine multi-modal inputs retrieved from RGB cameras, DVS, and LiDAR by creating some fusion layers in the network architecture.

\subsection{Bird's Eye View and LiDAR Representation}
By having a bird's eye view projection (BEVP), an autonomous driving agent will have a better scene understanding whether in the form of point-dot LS image or fully reconstructed BEVP image representations. In the field of BEVP, Reiher et al.\cite{birdview1} use four semantic segmentation images to construct BEVP. However, the model relies on other semantic segmentation models to provide four SS images. Thus, the entire process is not completed in one forward pass. With a similar concept, Palazzi et al.\cite{birdview4} develop a model that takes the front view RGB image along with its pre-predicted bounding boxes coordinate to estimate the bounding boxes on top view perspective. Both approaches may fail due to poor illumination problems (night and heavy rain) since they only use a monocular camera to provide the information. Another similar approach is conducted by Mani et al.\cite{birdview2} where a single model is used to estimate BEVP without any help from other models. However, the model cannot estimate another view since it takes the front RGB image as the only input.

Besides using several RGB cameras to capture multiple views of the surrounding, a $360^o$ LiDAR sensor can be used to collect numerous point clouds that contain the necessary information in a certain vertical field of view. Moreover, unlike RGB cameras, LiDAR provides useful data that is not affected by light illumination and weather conditions significantly. Currently, there have been plenty of studies that conduct research on processing LiDAR point clouds to fit the input of a deep learning model. Point cloud-based models\cite{lidar_pointnet}\cite{lidar_pointcnn} are known as the pioneers in taking the LiDAR point clouds directly, learning the feature, and predicting the label for each point. This mechanism is quite simple but the model tends to fail in capturing the local structure of an object. Then, in view-based models\cite{lidar_mvcnn}\cite{lidar_rotationnet}, LiDAR point clouds are projected into several 2D frames with multiple perspectives of views, then a simple convolution layer is used to process each frame. However, the number of possibilities of views can be large and lead to an expensive computational cost. Thus, an effective way to pre-process LiDAR point clouds is needed to reduce the computational load while still preserving useful information for the learning process.

In the BEVP task where surrounding objects are projected into top view perspective only, any unnecessary projection can be eliminated to save computational cost. Imad et al.\cite{birdview5} project raw LiDAR point clouds into a top view RGB image that has three channels so that a transfer learning method from various pre-trained models can be applied to perform the BEVP task. Although the heatmap coloring technique is used to differentiate the data, a lot of information still can be lost due to the limited scale. Then, an improved pre-processing approach is presented by Yang et al.\cite{pixor} where LiDAR point clouds are stored into a 3D tensor with the height information of the point cloud kept as the third dimension like channels in an RGB image. Thus a simple 2D convolution with a larger number of filters can be applied to process each channel. Using a proposed model called PIXOR, the pre-processed LiDAR data is used to perform top view object detection. This kind of data representation strategy is adopted by Zhang et al. to perform the top view LS task using the proposed model called PolarNet\cite{polarnet}. Finally, in line with the BEVP task, Chen et al.\cite{birdview3} develop a model that takes the top view LS image to perform the BEVP task. However, it means that another model is needed to support the LS task first.

\begin{figure*}
	\centering
	\includegraphics[width=\textwidth]{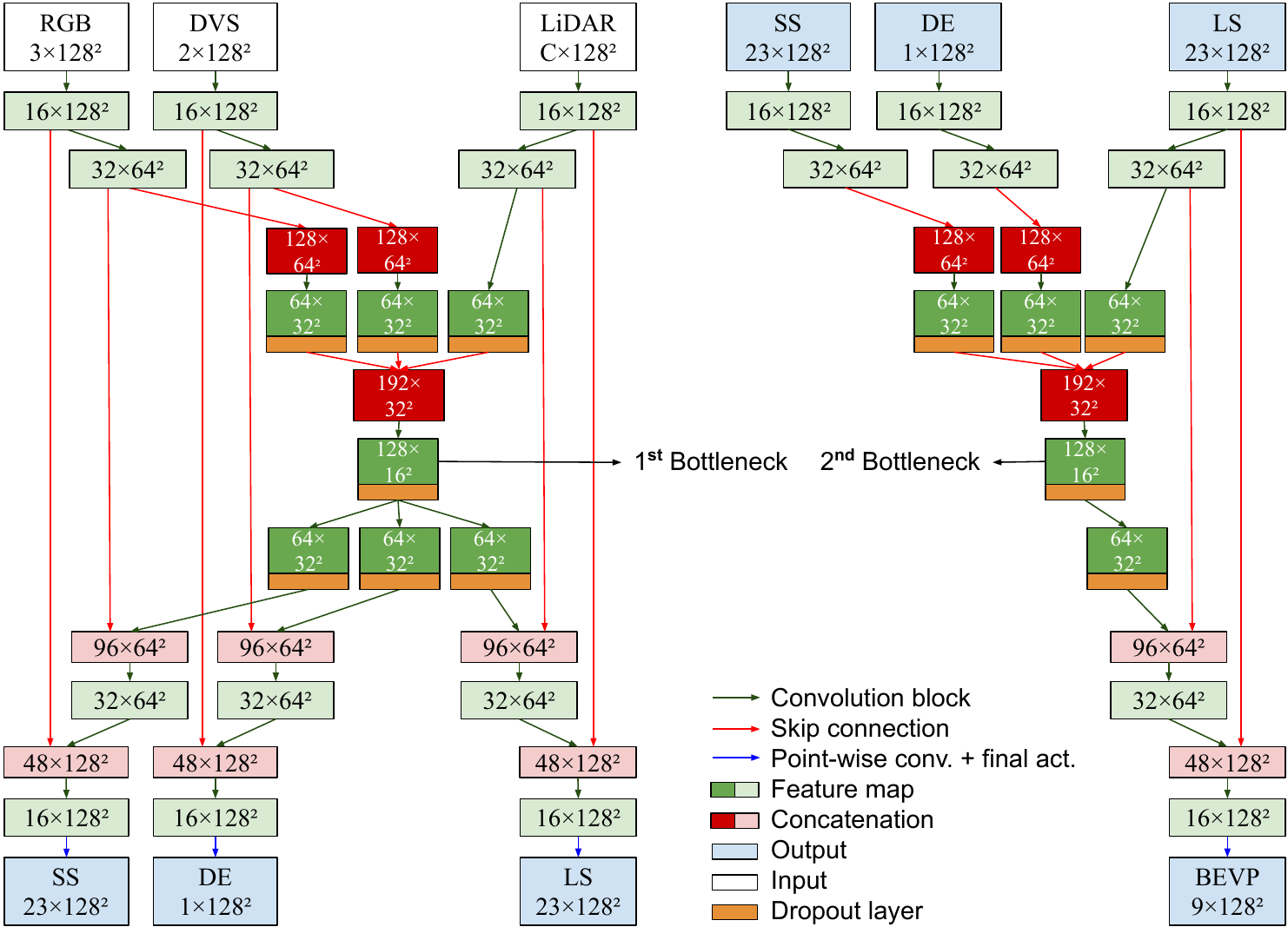}
	\caption{Network Architecture. To be noted, each view (Front, Left, Right, and Rear) on RGB, DVS, SS, and DE have its own encoder, while LiDAR and LS only have one encoder as there is only one view (Top). The text written inside each box is the tensor size in $C \times S^2$, where $C$ is the number of channels and $S^2$ is the spatial dimension (height $\times$ width). For LiDAR encoder, the network can take 1 or 15 layers of pre-processed LiDAR point clouds (see Subsection \ref{preprocessing}).}
	\label{fig:netarch}
\end{figure*}

In this research, we also utilize all front, left, right, rear images to support both LS and BEVP tasks as demonstrated by Reiher et al.\cite{birdview1}. Meanwhile, in pre-processing LiDAR point clouds, we combine two different techniques presented by Imad et al.\cite{birdview5} and Yang et al.\cite{pixor}. Therefore, we will have a 3D tensor that stores all point clouds into two forms of representations that contain more useful information.

\subsection{Multi-loss Weighting} \label{multiloss_ref}
A proper loss weighting strategy plays an important role in the training process of an MTL model, especially in tackling the imbalanced learning issue due to heterogeneous tasks with various loss functions. Cipolla et al.\cite{adaptloss:uncertainty} conduct research in multi-loss weighting on an MTL model that performs scene understandings such as semantic segmentation, instance segmentation, and depth regression simultaneously. By experimenting on an MTL dataset called Tiny Cityscapes\cite{cityscapes}, they show that homoscedastic task uncertainty is an effective way to perform loss weighting on several tasks. Meanwhile, a different approach is presented by Chen et al.\cite{adaptloss:gradnorm} where a loss weighting algorithm called Gradient Normalization (GradNorm) is proposed to control the training dynamics by manipulating the gradient during the training process. By adjusting the gradient signal, the learning conflict from different tasks can be minimized. 

In this research, we adopt GradNorm\cite{adaptloss:gradnorm} to deal with the imbalanced learning problem caused by plenty of tasks with different characteristics. We also do some modifications to the algorithm to meet our model needs.


\section{Methodology} \label{method}

In this section, we explain the details of our proposed methods which are inspired by some related works reviewed in Section \ref{relatedwork}. First, we describe the model architecture and the proper loss and metric formulation. Then, we explain how we develop the adaptive loss weighting algorithm to tackle the imbalanced learning issue. 

\subsection{Proposed Model}


As shown in Figure \ref{fig:netarch}, we use a common encoder-decoder style with a specific encoder and decoder for each input and output as demonstrated by Lv et al.\cite{MTL0}. Then, we add several skip connections to connect the feature maps on the encoder side with their symmetric feature maps on the decoder side inspired by the famous U-Net architecture\cite{unet}. This mechanism aims to enhance the model performance by leveraging combined feature maps on the bottleneck with the specifically extracted features from each decoder. Each RGB encoder is connected and concatenated to each semantic segmentation (SS) decoder that has the same view and spatial dimension. Rich color, shape, and much more extra information contained in the RGB image can be helpful for segmentation problems. Meanwhile, the DVS input is specifically used to support the depth estimation (DE) task by connecting and concatenating each pair of symmetric encoder-decoders in a similar way to the RGB-SS pair. DVS image can be helpful for estimation problems especially during poor illumination conditions (e.g. night) since it can contain contrast information even if there is only a small brightness change captured by the sensor. Then, we connect the encoder of pre-processed LiDAR point clouds to the LiDAR segmentation (LS) decoder inline with the LS encoder to the bird's eye view projection (BEVP) decoder as they have the same top perspective of view. Similar to Chen et al. \cite{birdview3}, our model performs BEVP by leveraging LS image. However, instead of taking the LS image directly as its input, we feed the model with the raw LiDAR point clouds that have been pre-processed to perform LS, then utilize the LS output to perform the BEVP task. Thus, there is no need to use another model to specifically support the LS task first. Our model is also leveraging 4 views of SS images as inspired by Reiher et al. \cite{birdview1} along with 4 views of DE images to support the LS encoder in performing BEVP. Finally, by following the intermediate fusion technique presented by Nie et al.\cite{SF_interfusion}, we create two bottlenecks in the form of convolution blocks to fuse and process multiple extracted feature maps from various input encoders. The $1^{st}$ bottleneck is used to store the extracted latent space from various inputs (RGB, DVS, and LiDAR) and used to perform SS, DE, and LS tasks. Meanwhile, the $2^{nd}$ bottleneck is used to store the extracted information from those tasks and perform BEVP as the final task. Hence, a compact network architecture that performs multiple tasks in one forward pass can be achieved.

The detailed explanation about the network architecture shown in Figure \ref{fig:netarch} is as follows. Each red line represents the skip connection that connects each pair of symmetric encoder-decoders followed by a concatenation process. The dark green line represent a common convolution block that consists of $2\times (3\times3$ convolution layer $+$ batch normalization\cite{batchnorm} $+$ ReLU activation\cite{relu}$)$ and followed with a $2\times2$ max-pooling layer for encoder path or $2\times2$ bilinear upsampling for decoder path. On the encoder path, the spatial dimension of the tensor is reduced by half while the number of feature maps in the channel axis is doubled each time it passes the convolution block. Meanwhile, on the decoder path, the spatial dimension is doubled while the number of feature maps is reduced by half gradually. Finally, the blue line represents a point-wise $1\times1$ convolution layer that reduces the channel so that the number of output elements will match with the number of channels of the ground truth. Then, a sigmoid layer is used to perform point-wise classification for SS, LS, and BEVP tasks and a ReLU layer for point-wise regression in a positive normalized range of 0 to 1 for the DE task. To prevent overfitting, we add a dropout layer\cite{dropout} with a drop rate of $p = 0.5$ on each convolution block at the center of the architecture. 


\subsection{Loss and Metric Formulation} \label{loss_metric}
To train the model and monitor its performance, loss and metric functions are needed to be formulated carefully. Loss functions are used to update the model weights while metric functions are used to monitor the model performance. Therefore, we use the metric scores to decide whether the training must be stopped or the learning rate should be reduced. To be noted, comparing loss values will not give a fair comparison due to different loss weights computed by the adaptive loss weighting algorithm. Hence, we use several metric scores as their calculation remains the same. As for the depth estimation (DE) loss ($\mathcal{L}_{DE}$), we calculate Huber loss as in (\ref{eq:huberloss1}).

\begin{equation} \label{eq:huberloss1}
	\mathcal{L}_{DE} = \frac{1}{V} \sum_{i = 1}^{V} \frac{1}{N} \sum_{j = 1}^{N} z_{ij} ,
\end{equation}

\noindent where $z_{ij}$ is given by (\ref{eq:huberloss2}).

\begin{equation} \label{eq:huberloss2}
	z_{ij} =
	\begin{cases}
			0.5 (\hat{y}_{ij} - y_{ij})^2 & \text{if $|\hat{y}_{ij} - y_{ij}| < \delta$} \\
			\delta(|\hat{y}_{ij} - y_{ij}| - 0.5\delta) & \text{otherwise} \\
	\end{cases}      
\end{equation}

We average the loss across all tensor elements $N$ on all views $V = 4$. The number of $N$ is also equal to the number of elements in pre-processed ground truth for DE task $I_{DE}$ (see Subsection \ref{preprocessing}). Then, $y_{ij}$ is the value of $j^{th}$ element of the ground truth $I_{DE}$ with view $i$, while $\hat{y}_{ij}$ is the predicted value of $j^{th}$ element of the predicted depth output with view $i$ after ReLU activation. Huber loss is widely used and suitable for the DE task as it takes the advantage of both mean squared error (MSE) and mean absolute error (MAE) based on the prediction results. We set $\delta = 0.5$ as the threshold for the Huber loss to start to curve like MSE if $|\hat{y}_{ij} - y_{ij}| < \delta$ or constantly have a large gradient which is the same as MAE if $|\hat{y}_{ij} - y_{ij}| \geq \delta$. Meanwhile, for the rest segmentation-related tasks, we use the combination of standard binary cross entropy (BCE) and Dice loss as in (\ref{eq:bcedice}) to calculate $\mathcal{L}_{SS}$, $\mathcal{L}_{LS}$, and $\mathcal{L}_{BEVP}$.

\begin{equation} \label{eq:bcedice}
\begin{split}
	\mathcal{L}_{\{SS,LS,BEVP\}} & =  \frac{1}{V} \sum_{i = 1}^{V} \bigg( \frac{1}{N} \sum_{j = 1}^{N} y_{ij} log(\hat{y}_{ij}) + (1-y_{ij})\\
	& log(1-\hat{y}_{ij}) \bigg) + \bigg( 1 - \frac{2|\hat{y}_i \cap y_i|}{|\hat{y}_i| + |y_i|} \bigg)
\end{split}
\end{equation}

\begin{figure*}
	\centering
	\includegraphics[width=\textwidth]{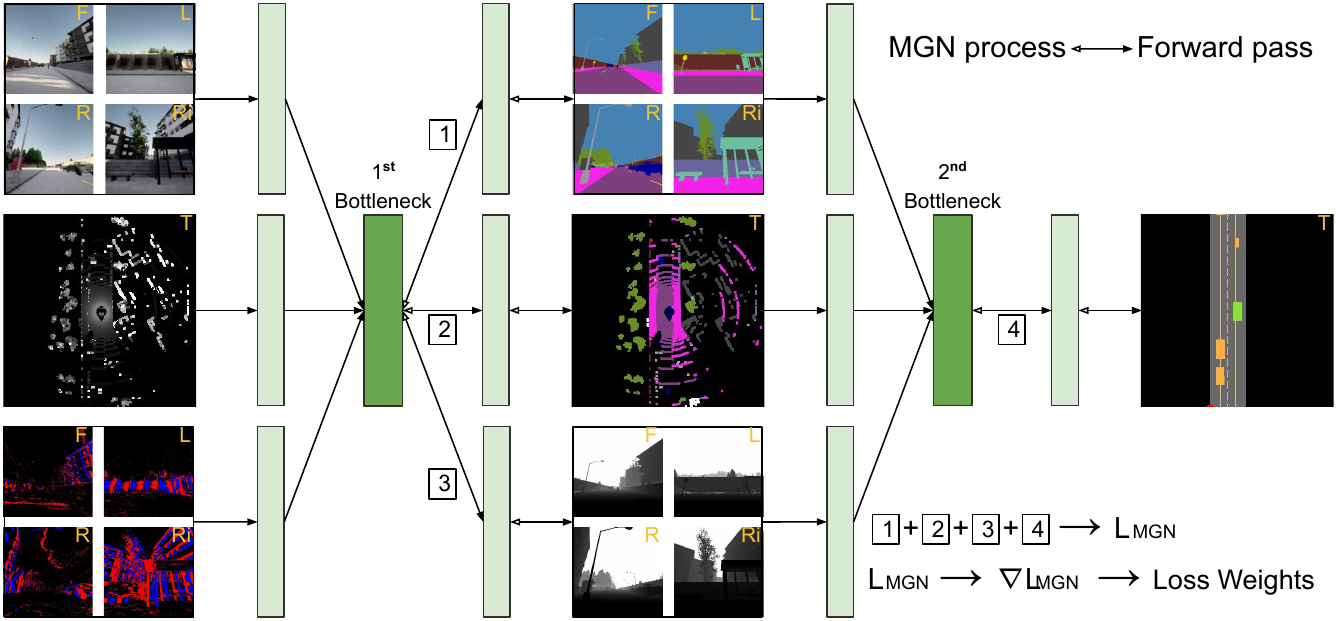}
	\caption{The overview of the MGN algorithm. The loss weight for each output on both SS and DE decoders is the same on each view since each SS loss and DE loss is averaged across all views.}
	\label{fig:gradnorm}
\end{figure*}

Similar to the Huber loss function, in the BCEDice loss function, the final loss calculation is also averaged across all tensor elements $N$ and all output views $V = 4$ for semantic segmentation (SS) task and $V = 1$ for LiDAR segmentation (LS) and bird's eye view projection (BEVP) tasks. Then, $y_i$ is the ground truth $I_{SS}$ or $I_{LS}$ or $I_{BEVP}$ with view $i$ and $\hat{y}_i$ is the predicted output of view $i$. Finally, the total loss can be calculated by multiplying each loss $\mathcal{L}_i$ with a loss weight $w_i$ and summing all of them. In addition, to prevent overfitting, a weight decay\cite{wdecay} with $\lambda = 0.0001$ is used to penalize model complexity by multiplying the sum-squared of model weights $\mathcal{W}$ and added to the total loss as in (\ref{eq:totalloss}).

\begin{equation} \label{eq:totalloss}
	\mathcal{L}_{total} = \lambda\Sigma\mathcal{W}^2 + \sum_{i = 1}^{T} w_i\mathcal{L}_i 
\end{equation}

To be noted, $\mathcal{L}_i$ is an element in a set of losses $\{ \mathcal{L}_{DE}, \mathcal{L}_{SS}, \mathcal{L}_{LS}, \mathcal{L}_{BEVP} \}$. Then, for the metric functions, we use MAE (\ref{eq:mae}) for depth estimation (DE) task and intersection over union (IoU) (\ref{eq:iou}) for segmentation-related tasks.

\begin{equation} \label{eq:mae}
	MAE_{DE} = \frac{1}{V} \sum_{i = 1}^{V} \frac{1}{N} \sum_{i = 1}^{N} |\hat{y}_{ij} - y_{ij}|
\end{equation}

\begin{equation} \label{eq:iou}
	IoU_{\{SS,LS,BEVP\}} = \frac{1}{V} \sum_{i = 1}^{V} \frac{|\hat{y}_i \cap y_i|}{|\hat{y}_i| \cup |y_i|}
\end{equation}

Finally, the total metric (TM) is calculated by summing all metric scores as formulated in (\ref{eq:totmetric}). To be noted, we use TM to determine the best model later as it represents overall model performance in all tasks. The total loss (\ref{eq:totalloss}) cannot be used for comparison as it is affected by multiplication of loss weights and weight decay which are varied amongst models. Then, in order to know the discrepancy between tasks and shows how balanced the performance across all tasks, we calculate the metric variance (MV) within $MAE_{DE}$, $1-IoU_{SS}$, $1-IoU_{LS}$, and $1-IoU_{BEVP}$ with (\ref{eq:metricvar}).

\begin{equation} \label{eq:totmetric}
\begin{split}
	TM & = MAE_{DE} + (1-IoU_{SS})\\
	& + (1-IoU_{LS}) + (1-IoU_{BEVP})
\end{split}
\end{equation}

\begin{equation} \label{eq:metricvar}
	MV = \frac{1}{T} \sum_{i = 1}^{T} \bigg(M_i - \frac{TM}{T}\bigg)^2 ,
\end{equation}

\noindent where $M_i$ is the metric score of task-$i$ and $\frac{TM}{T}$ is the mean of all metric scores with total tasks $T=4$. Keep in mind that the lower TM and MV scores mean the better and balanced the model performance.

\subsection{Adaptive Loss Weighting} \label{modgradnorm}

An adaptive loss weighting strategy can be used to deal with the imbalanced learning issue due to plenty of given tasks with different characteristics. In this research, we adopt the GradNorm algorithm\cite{adaptloss:gradnorm} and do some modifications to match our proposed model. The overall process of the modified GradNorm (MGN) algorithm can be seen in Figure \ref{fig:gradnorm}. Basically, the total loss for a multi-task model can be computed with (\ref{eq:staticlw}).

\begin{equation} \label{eq:staticlw}
  \mathcal{L}(t) = \sum_{i = 1}^{T} w_i(t) \mathcal{L}_i(t),
\end{equation}

\noindent where $\mathcal{L}_i$ is the loss function of task-$i$ from a $T$ number of tasks and a static loss $w_i$ is used to balance the learning process at training step $t$. Usually, the loss weights are tuned empirically which results in a huge computational cost to find the best set of loss weights. To address this issue, the GradNorm algorithm \cite{adaptloss:gradnorm} is invented to learn the loss weight $w_i$ by adjusting the gradient norms dynamically so that different tasks can be trained at similar rates. However, in their original paper, this algorithm is used to balance the learning process of the multi-task model of three tasks with only one input and one bottleneck of shared layers. Meanwhile, our model has four tasks with three data modalities, five different views as the input, and two bottlenecks of shared layers. Besides that, the loss weights are updated on each step and cause a huge computational load. We solve this issue by modifying the update process only at the last step $t=s$ for each epoch which means that there is only one update for one epoch. The number of maximum step $s$ is equal to the number of samples in the training set divided by the batch size. To be noted, the number of training samples on each dataset is different which means that the maximum step $s$ can be varied.


Before the loss weights are updated during the training phase on each epoch at step $t=s$, there are several quantities to be defined first with respect to the gradients that will be manipulated as follows:
\begin{itemize}
  \item Subsets of weights $W$ from entire model weights $\mathcal{W}$ where the algorithm will be applied. We pick 2 subsets of weights as there are 2 bottlenecks in the network architecture shown in Figure \ref{fig:netarch}. Mathematically, $W(s)$ selection is expressed as (\ref{eq:subsetW}).

  \begin{equation} \label{eq:subsetW}
	W(s) = \{W_1(s), W_2(s)\} \subset \mathcal{W}(s)
  \end{equation}
   
  Different from the original GradNorm paper\cite{adaptloss:gradnorm}, we pick $W_1(s)$ and $W_2(s)$ from the first layer of the $1^{st}$ and $2^{nd}$ bottleneck respectively. These layers are chosen since they have rich information of shared latent space from the concatenation of multiple feature maps.
  
  \item The $L_2$ norm of the gradient of the weighted single-task loss ($w_i(s)\mathcal{L}_i(s)$) with respect to the chosen subset of weights $W(s)$ that can be calculated with (\ref{eq:l2normgrad}).
  	\begin{equation} \label{eq:l2normgrad}
  		G_W^{(i)}(s) = \left\lVert\nabla_W (w_i(s) \mathcal{L}_i(s)) \right\rVert_2
	\end{equation}
  Based on the network architecture shown in Figure \ref{fig:netarch}, the gradient of the weighted single-task loss for BEVP is respected to $W(s) = W_2(s)$, while the others are respected to $W(s) = W_1(s)$. For further computation process, we need to compute $\overline{G}_W(s)$ which is the average of $G_W^{(i)}(s)$ across all tasks $T$ with (\ref{eq:avgl2normgrad}).
  	\begin{equation} \label{eq:avgl2normgrad}
  		\overline{G}_W(s) = \frac{1}{T} \sum_{i = 1}^{T} G_W^{(i)}(s)
  	\end{equation}
  
  \item The ratio between $\mathcal{L}_i$ at the last step $t=s$ and first step $t=0$ which can be computed with (\ref{eq:lossratio}).
  	\begin{equation} \label{eq:lossratio}
  		\bar{\mathcal{L}}_i(s) = \frac{\mathcal{L}_i(s)}{\mathcal{L}_i(0)}
  	\end{equation}
  Concisely, the loss ratio $\bar{\mathcal{L}}_i(s)$ is also a measure of the inverse training rate of task-$i$ where a lower ratio means a faster rate of learning task-$i$.
  
  \item The relative inverse training rate of task-$i$ which can be calculated with (\ref{eq:relativeinvrate}). This variable is used to balance gradients during the training process.
  	\begin{equation} \label{eq:relativeinvrate}
		r_i(s) = \frac{\bar{\mathcal{L}}_i(s)}{\frac{1}{T}\sum_{i = 1}^{T} \bar{\mathcal{L}}_i(s)}
	\end{equation}
	The higher relative inverse training rate $r_i(s)$ means the higher gradient magnitude for task-$i$ which results in the task being learned faster.
\end{itemize}

\begin{algorithm}[t] 
	\SetAlgoLined
	Initialize model weights $\mathcal{W}$ with kaiming init\cite{kaiming_init}\\
	Set initial loss weights $w_i(0) = 1 \forall_i$\\
	Set asymmetry alpha $\alpha = 1.5$\\
	Calculate maximum training step $s$*\\
	\For{training step t = 0 \KwTo s}{
		Standard forward pass:\\
		\begin{itemize}
			\item Input batch $x(t)$ and get prediction $\hat{y}(t)$\\
			\item Compute each single-task loss $\mathcal{L}_i(t)$\\
			\item Compute total loss $\mathcal{L}(t)$ with (\ref{eq:staticlw})\\
		\end{itemize}
		
		\uIf{t = 0}{
			Set initial task loss $\mathcal{L}_i(0) = \mathcal{L}_i(t)$\\
		}
		\uElseIf{t = s}{
			Pick $W(s)$ with (\ref{eq:subsetW})\\
			Compute $G_W^{(i)}(s)$ with (\ref{eq:l2normgrad}) for each task-$i$\\
			Compute $\overline{G}_W(s)$ with (\ref{eq:avgl2normgrad})\\
			Compute $\bar{\mathcal{L}}_i(s)$ with (\ref{eq:lossratio}) for each task-$i$\\
			Compute $r_i(s)$ with (\ref{eq:relativeinvrate}) for each task-$i$\\
			Compute $\mathcal{G}_W^{(i)}(s)$ with (\ref{eq:targetgrad})\\
			Compute $\mathcal{L}_{MGN}(s)$ with (\ref{eq:lgrad})\\
			Compute MGN gradients $\nabla w_i \mathcal{L}_{MGN}(s)$\\
			Update each $w_i(s)$ using $\nabla w_i \mathcal{L}_{MGN}(s)$\\
			Normalize new $w_i(s)$ with (\ref{eq:lwnormalization})\\
		}
		{\bf end}\\
		Standard backward pass:\\
		\begin{itemize}
			\item Compute gradients $\nabla_{\mathcal{W}} \mathcal{L}(t)$\\
			\item Update network weights $\mathcal{W}(t)$ using $\nabla_{\mathcal{W}} \mathcal{L}(t)$\\
		\end{itemize}
		
	}
	\caption{Training with MGN}
	\label{alg:gradnorm}
	\begin{tablenotes}
		\small
		\item *Maximum training step $s$ can be calculated by dividing the\\total training samples with batch size. It can be varied as the number of training samples is different on each dataset.
		
	\end{tablenotes}
\end{algorithm}

The detailed steps of the modified GradNorm (MGN) algorithm can be seen on Algorithm \ref{alg:gradnorm}. To be noted, there are only two training steps in one epoch to be considered for MGN computation which are the first step $t=0$ and the last step $t=s$. $\mathcal{L}_i(0)$ is very crucial especially at the first epoch of the training process. Thus, proper model weights $\mathcal{W}$ initialization and task loss $\mathcal{L}_i$ formulation need to be considered carefully. Furthermore, both depth estimation (DE) and semantic segmentation (SS) tasks have multiple inputs from 4 different views while LiDAR segmentation (LS) and bird's eye view projection (BEVP) tasks only have one input from a top perspective. Thus, the $\mathcal{L}_i$ for both DE and SS tasks are averaged across all views first before computing $\mathcal{L}_i(0)$ and $\mathcal{L}_i(s)$. This means that the loss weight $w_i$ for each DE task and SS task will be the same on any view. The MGN algorithm is deployed as a loss function that computes the MAE between the target and actual gradient norms as in (\ref{eq:lgrad}) for each task in every last step $t=s$ on each epoch.

\begin{equation} \label{eq:lgrad}
	\mathcal{L}_{MGN}(s) = \sum_{i = 1}^{T} \left\lvert \mathcal{G}_W^{(i)}(s) - G_W^{(i)}(s) \right\rvert ,
  \end{equation}

\noindent where the loss is summed across all tasks $T$ with the target gradient $\mathcal{G}_W^{(i)}(s)$ is given by (\ref{eq:targetgrad}).

\begin{equation} \label{eq:targetgrad}
\mathcal{G}_W^{(i)}(s) = \overline{G}_W(s) r_i(s)^\alpha
\end{equation}

We set the asymmetry $\alpha = 1.5$ as an additional parameter to control the balancing rate. The higher $\alpha$ value means the stronger balancing enforcement which is usually used if tasks are significantly different \cite{adaptloss:gradnorm}. Then, we use stochastic gradient descent (SGD) algorithm\cite{SGDm} to compute the gradients $\nabla w_i \mathcal{L}_{MGN}(s)$ and update the loss weights $w_i(s)$. We set the initial update rate $\eta_{MGN_0} = 0.1$ and reduce it by half until a minimum value of $\eta_{MGN_{min}} = 0.0001$ if there is no drop on total metric score in validation dataset in 4 epochs in a row. Finally, each loss weight $w_i(s)$ is normalized with (\ref{eq:lwnormalization}) so that the sum of all loss weights will always equal to $T$. 

\begin{equation} \label{eq:lwnormalization}
  w_{i_{new}}(s) = \frac{w_i(s)}{\sum_{i = 1}^{T} w_i(s)} T
\end{equation}

\section{Experiment Setup} \label{expsetup}

In this section, we describe the dataset used for experiments which consist of 3 simulation datasets and 1 real-world dataset. Then, the pre-processing steps are explained to understand the data representation. We also provide a brief explanation about the training configuration.

\subsection{Simulation Data} \label{dataset}

\begin{figure}[!t]
  \centering
  \includegraphics[width=\linewidth]{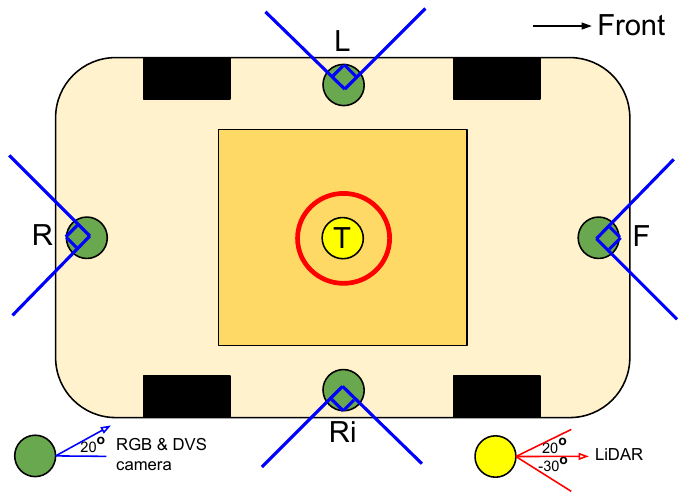}
  \caption{Sensors Placement. Both RGB and DVS cameras are placed at four different positions, while the $360^o$ LiDAR sensor is placed at the top of the ego vehicle. Each sensor has a specific horizontal and vertical field of view.}
  \label{fig:sensorplace}
\end{figure}


We use CARLA simulator\cite{carla} to generate simulation datasets to train our model. We collect a large amount of data composed of RGB images, DVS arrays, and LiDAR point clouds as the inputs and semantic segmentation (SS) images, depth estimation (DE) images, LiDAR segmentation (LS) images, and bird's eye view projection (BEVP) images as the outputs. In this research, we create three different datasets named dataset A, B, and C for the experiment and strengthen our justification. In dataset A, we gather the simulation data from map 'town01' as the training set and 'town02' for both validation and test sets. Then, in dataset B, we collect the training set from 'town02' and the rest validation and test sets from 'town01'. Meanwhile, in dataset C, we generate all simulation data from all maps ('town01' to 'town05') for training, validation, and testing sets. Each map has different characteristics and contains various objects.

\begin{table}[t] 
	\caption{Data Generation Setting}
	\begin{center}
		\begin{tabular}{|p{0.35\linewidth}|p{0.55\linewidth}|}
			\hline
			{\bf Parameter} & {\bf Configuration}\\
			\hline
			Train : Val : Test ratio & 3 : 1 : 1\\
			\hline
			Total data & 2000 (set A and B), 10000 (set C)\\
			\hline
			Maps used & 2 (set A and B), 5 (set C)\\
			\hline
			Simulation time & Morning, noon, evening, and night\\
			\hline
			Weather & Sunny, rainy, cloudy, and foggy\\
			\hline
			Non-player characters & Other vehicles (truck, car, bicycle, motorbike) and pedestrians\\
			\hline
			Object class for SS and LS (23 classes) & Unlabeled, building, fence, other, pedestrian, pole, road lane, road, side walk, vegetation, other vehicles, wall, traffic sign, sky, ground, bridge, rail track, guard rail, traffic light, static object, dynamic object, water, terrain\\
			\hline
			Object class for BEVP (9 classes) & Road, road lane, road centerline, other vehicles, ego vehicle, green traffic light, yellow traffic light, red traffic light, pedestrian\\
			\hline
			CARLA version & 0.9.10.1\\
			\hline
		\end{tabular}
	\end{center}
	\label{tab:data_info}
\end{table}

To obtain more information about the surroundings, several sensors are mounted on the ego vehicle as shown in Figure \ref{fig:sensorplace}. We place RGB and DVS cameras at four positions on the ego vehicle, which are front (F), left (L), right (Ri), and rear (R). Each camera has a $90^o$ horizontal and vertical field of view, $20^o$ upward rotation, and original resolution of $H \times W = 128 \times 128$. Then, a $360^o$ LiDAR sensor with 64 lasers and 32 meters of the maximum range is placed at the top (T) of the vehicle. The LiDAR lasers are vertically spread between the range of $-30^o$ to $20^o$ from the horizontal line. The same configuration is also applied to get the ground truth data for each task. During the data gathering process, we create a realistic condition by changing the weather dynamically where the environment can be sunny, rainy, foggy, morning, noon, evening, and night. Each condition also varies on a scale of 0 to 100\% and can be combined. In addition, we also spawn non-player characters such as pedestrians and other vehicles to mimic the real situation on the road. The detailed information of the data generation setting can be seen in Table \ref{tab:data_info}.

\subsection{Real-world Scenario} \label{realworld_exp}

\begin{figure*}
	\centering
	\includegraphics[width=\linewidth]{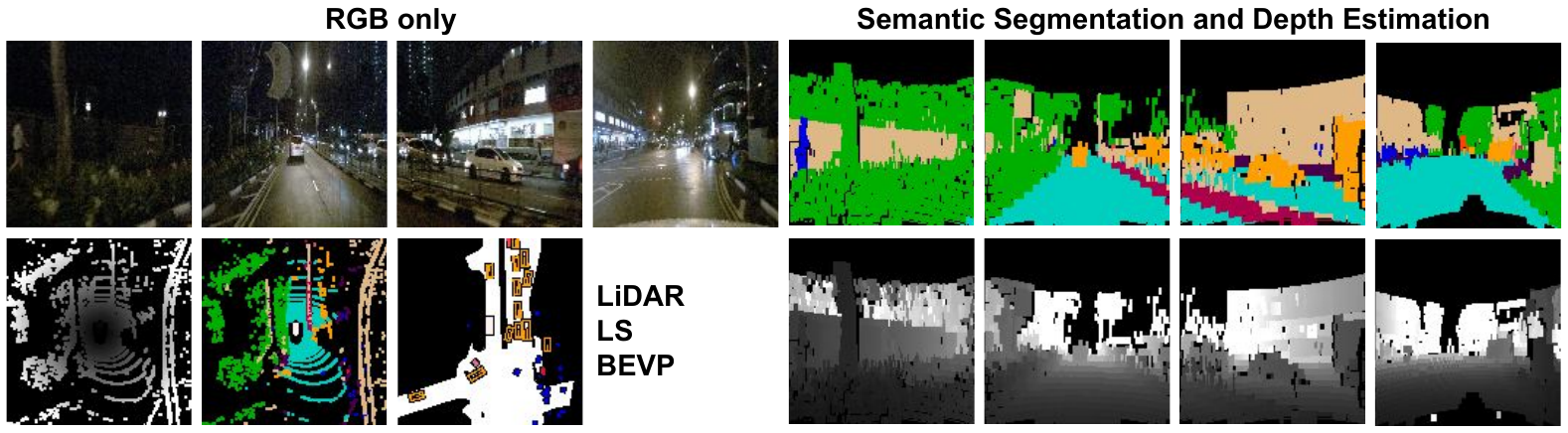}
	\caption{A set of pre-processed nuScenes-lidarseg samples. From left to right views: L-F-Ri-R while LiDAR, LS, and BEVP only have a top view. There are no DVS images for model inputs so all DE decoders will retrieve extracted RGB feature maps only. Since there are no ground truths given for DE and SS tasks, LiDAR point cloud's associated class and distance information are used to create the ground truth for both tasks.}
	\label{fig:nuscene_sample}
\end{figure*}

To illustrate how our model can be deployed in a real-world scenario, we also use nuScenes-lidarseg dataset\cite{nuscene} as the fourth dataset in our experiment. However, this dataset has a different sensor configuration and is not providing DVS images and ground truth for both semantic segmentation (SS) and depth estimation (DE) tasks. Thus, we consider modifying and pre-processing the dataset to meet the model needs. First, we use front-left and front-right images as the replacement for left and right images which are not provided in nuScenes-lidarseg. As there are no DVS images for the model inputs, we remove all DVS encoders in the network architecture and branch each RGB encoder to support both SS and DE decoders. Therefore, the model will take RGB inputs only to perform DE and SS tasks. Moreover, this dataset does not come with the ground truth for SS and DE tasks. Thus, we use the provided LiDAR point clouds associated class and distance information to create ground truths for SS and DE tasks. We create SS and DE ground truths using the point clouds that are shown on each camera's perspective of view. Concisely, we plot each point cloud's associated class data as the SS ground truth and distance data as the DE ground truth. Then, to fill the gap between plotted frame's pixels, we give neighboring pixels the same class or value as the filled pixel. With this mechanism, we can obtain nearly similar ground truths as retrieved from the CARLA simulator. To be noted, we also resize all images to have a spatial dimension of $H \times W = 128 \times 128$ which is the same as in datasets A, B, and C. Thus, there is no need to make any further modifications to the model input size. Finally, since nuScenes-lidarseg has 32 possible object classes to be recognized, therefore, the number of channels of semantic segmentation (SS), LiDAR segmentation (LS), and bird's eye view projection (BEVP) outputs become $C = 32$. A set of nuScenes-lidarseg samples can be seen in Figure \ref{fig:nuscene_sample}.

Originally, the nuScenes-lidarseg dataset has 1000 driving scenes obtained from Boston and Singapore that have dense traffic and challenging situations. However, we only use the original 'trainval' set (850 scenes) for our experiment. Meanwhile, the original 'test' set (150 scenes) is excluded since the ground truth is not publicly available. Hence, we cannot measure the performance of the model. The nuScenes-lidarseg's 'trainval' set has a total sample of 34149 from 850 scenes that are different from one another. We divide the original 'trainval' set into train, validation, and test sets with the ratio of 3:1:1 (the same as in datasets A, B, and C) based on the number of scenes so that the sample will be completely different on each set. Thus, there are 510 scenes (20418 samples) for training, 170 scenes (6873 samples) for validation, and 170 scenes (6858 samples) for testing. 

\begin{figure*}
	\centering
	\includegraphics[width=\linewidth]{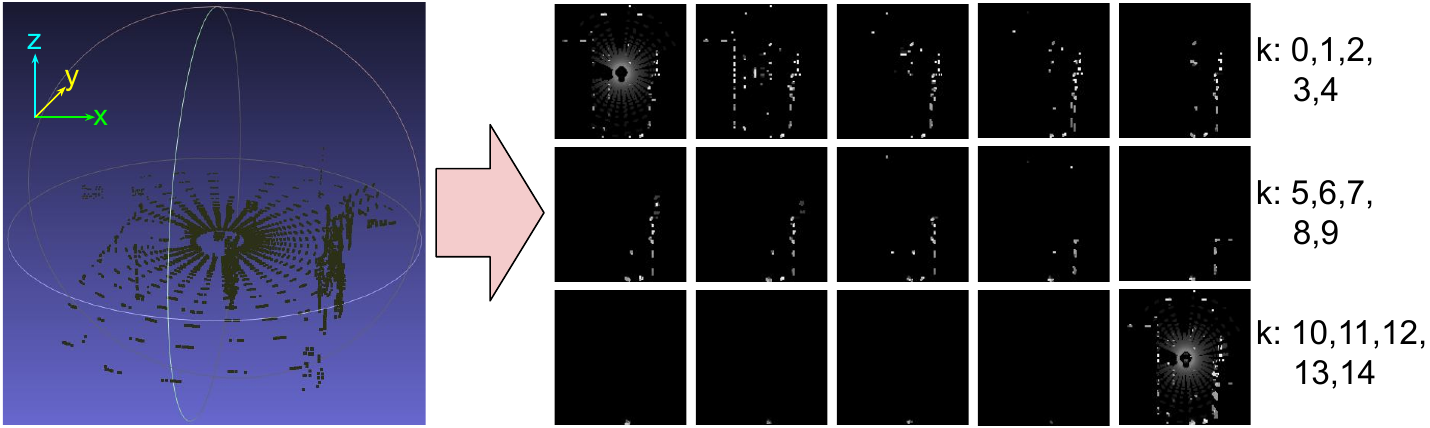}
	\caption{LiDAR point clouds pre-processing. All point clouds are mapped into a tensor $I_{LID} \in \{0,...,1\}^{15 \times 128 \times 128}$ where each layer holds the height information that spreads within an interval of -2 meter (lowest point) to 11 meters (highest point).}
	\label{fig:lidpreprocess}
\end{figure*}

\subsection{Data Pre-processing and Representation} \label{preprocessing}



RGB image is originally retrieved as $I_{RGB} \in \{0,...,255\}^{3 \times 128 \times 128}$ representing the set of 8-bit pixel value in the form of RGB channel ($C$) $\times$ height ($H$) $\times$ width ($W$). Different from $I_{RGB}$, the DVS array is retrieved as $A_{DVS} \in \mathbb{R}^{N \times 4}$ where ${N}$ is the total number of pixels that are considered to have a brightness change in one simulation step and $\mathbb{R}$ is an array with four elements consisted of timestamp, pixel's x-coordinate, pixel's y-coordinate, and pixel's polarization. The pixel's polarization can be positive or negative depending on the brightness change. Meanwhile, a set of LiDAR point clouds is retrieved as $A_{LID} \in \mathbb{R}^{M \times 4}$ where ${M}$ is the total number of point clouds retrieved in one simulation step and $\mathbb{R}$ is an array with four elements composed of point's x-coordinate, point's y-coordinate, point's z-coordinate, and cos of incident angle (cos($\theta$)).

As for the model input, we normalize all RGB images in a scale of 0 to 1 expressed as $I_{RGB} \in \{0,...,1\}^{3 \times 128 \times 128}$. Since the model already takes multiple inputs of views, there is no need to feed the model with a bigger input size. Thus, it can reduce the computational load during both training and validation processes. To meet the input shape of the model, both $A_{DVS}$ and $A_{LID}$ need to be pre-processed first due to the possibility of numerous ${N}$ and ${M}$ which are affected by the simulation condition at a time. Thus, to deal with this issue, we pre-process both $A_{DVS}$ and $A_{LID}$ to become a 3D tensor with a fixed shape. With a maximum x,y coordinate range of $(127,127)$, we project $A_{DVS}$ into $I_{DVS} \in \{0,1\}^{2 \times 128 \times 128}$ with the x,y-coordinate takes place on the spatial dimension $(H \times W = 128 \times 128)$ and the polarization takes place on the channel dimension $C = 2$. In the channel axis, the positive polarization takes place on the first channel while the negative polarization takes place on the second channel of $I_{DVS}$. We convert the polarization status into a value of 0 or 1 with (\ref{eq:convertdvs}). Meanwhile, the timestamp attribute in $A_{DVS}$ is used to synchronize with other sensor data.

\begin{equation} \label{eq:convertdvs}
  I_{DVS_{ij}} = 
  \begin{cases}
          1 & \text{if there is polarization record in $A_{DVS_{xy}}$} \\
          0 & \text{otherwise},
  \end{cases}      
\end{equation}

\noindent where $I_{DVS_{ij}}$ is the i,j-coordinate derived from $A_{DVS}$'s x,y-coordinate. Meanwhile, for the LiDAR point clouds, we propose two different techniques to pre-process the LiDAR point clouds and perform an ablation study to understand their influence. The first method ignores $A_{LID}$'s z-coordinate and projects the LiDAR point clouds into a tensor with top view perspective $I_{LID} \in \{0,...,1\}^{1 \times 128 \times 128}$ with (\ref{eq:lidarproj1}). This kind of input representation is similar to Imad et al. \cite{birdview5}, however, instead of using heatmap color (in 3 channel RGB), we only use one channel to store the cosine of the incident angle (cos($\theta$)) of each point cloud. Therefore, a newly formed tensor $I_{LID}$ will likely form a heat map-like gray image. 

\begin{equation} \label{eq:lidarproj1}
  I_{LID_{ij}} = 
  \begin{cases}
          cos(\theta) & \text{if there is a point record in $A_{LID_{xy}}$} \\
          0 & \text{otherwise}
  \end{cases}      
\end{equation}

To be noted, the $I_{LID}$'s i,j-coordinate has been shifted and scaled from the original $A_{LID}$'s x,y-coordinate. The center $(x,y) = (0,0)$ of $A_{LID}$ is at the top of the ego vehicle and the range of x and y-axis are -32 to 32 meters (the maximum range of LiDAR). Therefore, we need to shift and scale the $I_{LID}$'s i,j-coordinate so that the center is at $(i,j) = (64,64)$ and the minimum and maximum coordinate are at $(i,j) = (0,0)$ and $(i,j) = (127,127)$ respectively. The shifting and scaling process of the i,j-coordinate from the original x,y-coordinate can be done with (\ref{eq:lidarcoordinatex}) and (\ref{eq:lidarcoordinatey}) respectively.

\begin{equation} \label{eq:lidarcoordinatex}
  I_{LID_i} = \left\lfloor \frac{A_{LID_x} - (-32)}{32 - (-32)} \times 127 \right\rceil
\end{equation}

\begin{equation} \label{eq:lidarcoordinatey}
  I_{LID_j} = \left\lfloor \frac{A_{LID_y} - (-32)}{32 - (-32)} \times 127 \right\rceil
\end{equation}

Both $I_{LID_i}$ and $I_{LID_j}$ represent the $I_{LID}$'s i,j-coordinate while $A_{LID_x}$ and $A_{LID_y}$ represent $A_{LID}$'s x,y-coordinate. Meanwhile, 127 is set to be the highest point of $I_{LID}$'s i,j-coordinate. Then, we also give a value to the nearest pixels from $I_{LID_{ij}}$ as the same as the pixel's value of $I_{LID_{ij}}$ itself. Thus, the pre-processed $I_{LID}$ will have a better area coverage from the top perspective of view. However, the first method can lose points that have the same $A_{LID}$'s x,y-coordinate but with a lower $A_{LID}$'s z-coordinate since the method only stores one point with the highest $A_{LID}$'s z-coordinate.

In the second LiDAR pre-processing method, we adopt the LiDAR pre-processing technique presented by Yang et al.\cite{pixor} that takes the $A_{LID}$'s z-coordinate into account. Then, we stack the pre-processed point clouds with the data from the first method to provide more rich information. The visualization of this method can be seen in Figure \ref{fig:lidpreprocess}. Concisely, the second method projects the LiDAR point clouds into a 3D tensor $I_{LID} \in \{0,...,1\}^{15 \times 128 \times 128}$. Here, we set the number of channels $n(k) = 15$ where $k \in \{0,...,14\}$ based on the vertical field of view and the maximum range of the LiDAR sensor. As can be seen in Figure \ref{fig:sensorplace}, the sensor has a $30^o$ view below and $20^o$ view above the horizon line. Since the sensor is placed on the top of the ego vehicle which is 2 meters from the ground, then the lowest point of the point cloud is equal to -2. Meanwhile, the highest point of the point cloud can be calculated with $\lceil sin(20) \times 32\rceil = 11$. Therefore, we set $n(k) = 15$ where the first 14 channels ($k \in \{0,...,13\}$) are used to store point clouds based on their height defined by $A_{LID}$'s z-coordinate which are spreading from the lowest point of -2 meter to the highest point of 11 meters. Then, the last channel ($k = 14$) is used to store all flattened point clouds from the first method. We map the $A_{LID}$'s z-coordinate into k channels with (\ref{eq:lidarcoordinatez}).

\begin{equation} \label{eq:lidarcoordinatez}
  I_{LID_k} = \left\lfloor \frac{A_{LID_z} - (-2)}{11 - (-2)} \times 13 \right\rceil ,
\end{equation}

\noindent where $I_{LID_k}$ is the $I_{LID}$'s k-coordinate (channels) and $A_{LID_z}$ is the original $A_{LID}$'s z-coordinate. The multiplier of 13 is used to ensure that there are no point clouds stored in the last channel ($k = 14$) as it has been reserved to store all flattened point clouds from the first method. The maximum and minimum values of $A_{LID_z}$ are set to 11 and -2 for all recorded point clouds so that the scale for all pre-processed $I_{LID}$ will be the same. Thus, the network can easily learn to segment objects based on their height even if the maximum point of each recorded point cloud is different. Finally, similar to the first method, we also give the same value to the nearest pixels around $I_{LID_{ijk}}$. To be noted, if there are 2 points or more with different $A_{LID}$'s x,y,z-coordinate but have the same $I_{LID}$'s i,j,k-coordinate after pre-processing, only the point that has the highest $A_{LID}$'s z-coordinate that can take place on the $I_{LID}$'s i,j,k-coordinate to prevent multiple data points stored in one coordinate. Therefore, having a large number of k channels would be better since there is more space to store point clouds. However, processing a larger input would also cost more computational time.


On the output side, we read the depth estimation (DE) ground truth as a tensor $I_{DE} \in \{0,...,1\}^{1 \times 128 \times 128}$. Thus, the output layer of the model for the DE task will only have 1 channel with the spatial dimension of $(128 \times 128)$ that predicts normalized depth value within the range of 0 to 1. With this mechanism, we can use simple ReLU activation for the final output layer of the DE decoder. Meanwhile, the original semantic segmentation (SS), LiDAR segmentation (LS), and bird's eye view projection (BEVP) ground truths are retrieved as $I_{\{SS,LS,BEVP\}} \in \{0,...,255\}^{3 \times 128 \times 128}$ which are following the color palette in Cityscapes dataset\cite{cityscapes}. To meet the needs of the network architecture, especially on its output layers for segmentation-related tasks (SS, LS, and BEVP), we perform one hot encoding process to convert the 8-bit RGB representation. As a result, each ground truth become $I_{SS} \in \{0,1\}^{23 \times 128 \times 128}$, $I_{LS} \in \{0,1\}^{23 \times 128 \times 128}$, and $I_{BEVP} \in \{0,1\}^{9 \times 128 \times 128}$. Therefore, a sigmoid activation can be used at the output layer of SS, LS, and BEVP decoders. The number of classes in the CARLA simulation dataset is 9 for the BEVP task and 23 for both SS and LS tasks as mentioned in Table \ref{tab:data_info}. Meanwhile, in the real-world nuScenes-lidarseg dataset, the number of classes is 32 for all SS, LS, and BEVP tasks as mentioned in Subsection \ref{realworld_exp}. Hence, the channel axis $C$ of $I_{\{SS,LS,BEVP\}}$ has also become 32.

\subsection{Training Configuration} \label{train_config}
We use two GPUs, the NVIDIA RTX 2080 super and GTX 1080 Ti separately to train the model along with its variation described in Section \ref{results}. In this research, we develop the model entirely from scratch using PyTorch\cite{torch}. We do not use any pre-trained network to perform transfer learning nor fine-tuning. As mentioned in Subsection \ref{modgradnorm}, weights initialization can be crucial as it affects $\mathcal{L}_i(0)$ especially at the early epoch of the training process. Therefore, the kaiming initialization strategy\cite{kaiming_init} is used to initialize the entire model weights $\mathcal{W}$. Then, a small batch size of 6 is enough since the model already takes multiple views of inputs. Similar to the loss weights updates, we use SGD\cite{SGDm} with momentum $\mu = 0.9$ to update the model weights during the training process. We set the initial learning rate $\eta_0 = 0.1$ and reduce it by half gradually until $\eta_{min} = 0.00001$ if there is no drop on the validation total metric (TM) score in 4 epochs in a row. We also stop the training process automatically if there is no drop in validation TM score in 25 epochs in a row.

\section{Result and Discussion} \label{results}

To evaluate our proposed methods, ablation and comparative study are conducted by comparing all model variants along with the combination of single-task model and multi-task model for all given tasks. For depth estimation (DE) and semantic segmentation (SS) tasks, we compare our models with the multi-task GradNorm model \cite{adaptloss:gradnorm}. In the LiDAR segmentation (LS) task, we compare our model with PolarNet\cite{polarnet} that performs the same top view LiDAR segmentation. Finally, for the bird's eye view projection (BEVP) task, we replicate the works by Chen et al. \cite{birdview3} and perform some modifications in their model's final output layer to be a point-wise convolution layer for one-hot encoded prediction so that we can calculate the IoU and perform a fair comparison. The best model is defined by the lowest total metric score as formulated with (\ref{eq:totmetric}). Moreover, as mentioned in Section \ref{expsetup}, we compare all models on 3 simulation datasets generated by CARLA simulator\cite{carla} and one real-world dataset from nuScenes-lidarseg\cite{nuscene}. Concisely, there are 3 points that will be disclosed in this research as follows:
\begin{itemize}
  \item The influence of providing 15-layer LiDAR data into the model. In this experiment, we compare two models where one model takes 1 layer of LiDAR data (1L) and the other takes 15 layers of LiDAR data (15L). Then, we observe its influence based on the TM score.
  \item The influence of using the MGN algorithm on the MTL model during the training process. By using the MGN algorithm, the model is expected to have better performance as the imbalanced learning problem will be solved by giving appropriate weight to each loss function. Therefore, to understand its effectiveness, a comparative study is conducted on the model with adaptive loss weights (15L+MGN) and the model with static loss weights (15L with $w_i = 1 \forall_i$). We also provide a separate subsection to discuss the behavior of this algorithm.
  \item A comparative study with the combination of some recent models. We compare all of our model variants with 2 single-task models and 1 multi-task model which are PolarNet\cite{polarnet} for LS, Chen et al \cite{birdview3} for BEVP, and GradNorm model\cite{adaptloss:gradnorm} for multi-task DE and SS. Besides calculating all metric scores, we also compute the number of model parameters, model size, GPU memory usage, and inference speed to justify the model efficiency.
  \end{itemize}




\subsection{1 Layer vs 15 Layers of LiDAR Representation} \label{1vs15}

\begin{table*}
	\begin{center}
				\begin{tabularx}{\textwidth}{*8c}
					\toprule
					Dataset & Model & $MAE_{DE}$ & $IoU_{SS}$ & $IoU_{LS}$ & $IoU_{BEVP}$ & $TM$ & $MV$\\
					\midrule
					& Chen et al.\cite{birdview3} & - & - & - & {\bf 0.654 $\pm <$0.001} & &  \\
					& PolarNet\cite{polarnet} & - & - & {\bf 0.483 $\pm <$0.001} & - & 1.430 & {\bf 0.024}\\
					Set & GradNorm$^{\dagger}$\cite{adaptloss:gradnorm} & 0.113 $\pm <$0.001 & 0.546 $\pm$ 0.002 & - & - & & \\
					A & 1L & 0.090 $\pm <$0.001 & 0.619 $\pm$ 0.001 & 0.406 $\pm <$0.001 & 0.613 $\pm <$0.001 & 1.452 & 0.032\\
					& 15L & {\bf 0.083 $\pm <$0.001} & {\bf 0.636 $\pm$ 0.002} & 0.424 $\pm <$0.001 & 0.575 $\pm <$0.001 & 1.448 & 0.032\\
					& {\bf 15L+MGN} & 0.084 $\pm <$0.001 & 0.627 $\pm$ 0.002 & 0.470 $\pm <$0.001 & 0.594 $\pm <$0.001 & {\bf 1.393} & 0.027\\
		
					\hdashline
					& Chen et al.\cite{birdview3} & - & - & - & 0.567 $\pm$ 0.003 &  & \\
					& PolarNet\cite{polarnet} & - & - & {\bf 0.723 $\pm <$0.001} & - & 1.160 & 0.016\\
					Set & GradNorm$^{\dagger}$\cite{adaptloss:gradnorm} & {\bf 0.095 $\pm <$0.001} & 0.645 $\pm$ 0.003 & - & - &  & \\
					B & 1L & 0.096 $\pm <$0.001 & 0.675 $\pm <$0.001 & 0.621 $\pm$ 0.001 & 0.589 $\pm$ 0.002 & 1.211 & 0.015\\
					& 15L & {\bf 0.095 $\pm <$0.001} & {\bf 0.682 $\pm <$0.001} & 0.680 $\pm <$0.001 & 0.603 $\pm$ 0.001 & 1.131 & 0.013\\
					& {\bf 15L+MGN} & 0.099 $\pm <$0.001 & 0.679 $\pm <$0.001 & 0.704 $\pm <$0.001 & {\bf 0.630 $\pm$ 0.002} & {\bf 1.086} & {\bf 0.011}\\
		
		
					\hdashline
					& {\bf Chen et al.\cite{birdview3}} & - & - & - & {\bf 0.637 $\pm <$0.001} &  & \\
					& {\bf PolarNet\cite{polarnet}} & - & - & {\bf 0.735 $\pm$ 0.001} & - & {\bf 0.976} & {\bf 0.013}\\
					Set & {\bf GradNorm$^{\dagger}$\cite{adaptloss:gradnorm}} & {\bf 0.055 $\pm <$0.001} & 0.706 $\pm <$0.001 & - & - & & \\
					C & 1L & 0.069 $\pm <$0.001 & 0.751 $\pm <$0.001 & 0.573 $\pm$ 0.001 & 0.606 $\pm <$0.001 & 1.138 & 0.020\\
					& 15L & 0.062 $\pm <$0.001 & {\bf 0.765 $\pm <$0.001} & 0.645 $\pm$ 0.001 & 0.602 $\pm <$0.001 & 1.050 & 0.017\\
					& 15L+MGN & 0.063 $\pm <$0.001 & 0.756 $\pm <$0.001 & 0.678 $\pm <$0.001 & 0.630 $\pm <$0.001 & 0.999 & 0.014\\
		
		
					\hdashline
					& Chen et al.\cite{birdview3} & - & - & - & 0.788 $\pm <$0.001 &  & \\
					& PolarNet\cite{polarnet} & - & - & {\bf 0.696 $\pm <$0.001} & - & 1.124 & 0.020\\
					nuScenes & GradNorm$^{\dagger}$\cite{adaptloss:gradnorm} & {\bf 0.112 $\pm <$0.001} & 0.504 $\pm$ 0.005 & - & - & &  \\
					-lidarseg & 1L* & 0.119 $\pm <$0.001 & 0.527 $\pm$ 0.009 & 0.597 $\pm$ 0.001 & 0.804 $\pm <$0.001 & 1.191 & 0.021\\
					& 15L* & 0.123 $\pm <$0.001 & {\bf 0.538 $\pm$ 0.007} & 0.682 $\pm <$0.001 & 0.824 $\pm <$0.001 & 1.079 & {\bf 0.017}\\
					& {\bf 15L+MGN*} & 0.123 $\pm <$0.001 & 0.536 $\pm$ 0.008 & 0.685 $\pm <$0.001 & {\bf 0.833 $\pm <$0.001} & {\bf 1.069} & {\bf 0.017}\\
					\bottomrule 
				\end{tabularx}
	\end{center}
	\caption{Performance Comparison on Test Sets}
	\label{tab:testcompare}
	\begin{tablenotes}
		\small
		\item $^{\dagger}$Scores are averaged across all views performed by 4 independent GradNorm models.
		\item *The model only takes extracted feature maps from RGB encoders to perform DE and SS as there is no DVS data in nuScenes-lidarseg.\\
		To be noted, the higher IoU and the lower MAE, total metric (TM), and metric variance (MV) scores mean the better the model. Keep in mind that the TM score is used to determine the best model as it represents overall performance on all tasks. Meanwhile, the uncertainty on each metric score is calculated by computing the variance across all inference results on each test set.
	\end{tablenotes}
\end{table*}

LiDAR point clouds contain a z-coordinate that represents the height of the object captured by the LiDAR lasers. The idea of our second LiDAR pre-processing method is to differentiate the object based on height data so that the model can leverage this useful information during the training process. As shown in Figure \ref{fig:lidpreprocess}, each layer contains a specific object based on its height. For example, the lower layer holds objects which are mostly on the ground such as roads, sidewalks, etc. Then, the middle layer holds other vehicles, pedestrians, etc. Then, the upper layer holds tall objects such as buildings, trees, etc. Finally, stacking all point clouds pre-processed by the first method into the last layer will provide more information. 

As shown in Table \ref{tab:testcompare}, the model that takes 15-layer LiDAR data (15L) has a better performance compared to the model that takes 1 layer only (1L). The comparison between both models is consistent where the 15L model has a lower total metric (TM) score than the 1L model on all test sets. The TM score get lowered from 1.452 to 1.448 (set A), 1.211 to 1.131 (set B), 1.138 to 1.050 (set C), and 1.191 to 1.079 (nuScenes-lidarseg). Intuitively, adding more layers of information will boost the LS performance as it has inline skip connections from LiDAR encoder to LS decoder. This is proven by comparing the $IoU_{LS}$ score where the 15L model has a higher score than the 1L model on all test sets. However, in the BEVP task, both model variants are comparable to each other as the 15L model has higher $IoU_{BEVP}$ scores on dataset B and nuScenes-lidarseg but has lower scores on dataset A and set C. Then, the other interesting thing is the result of DE and SS tasks. Based on $MAE_{DE}$ and $IoU_{SS}$ scores, we found that adding more LiDAR layers is somehow improving DE and SS performance. Consistently, the 15L model has higher $IoU_{SS}$ and lower $MAE_{DE}$ than the 1L model on all simulation datasets, and only the DE performance is degraded on nuScenes-lidarseg. As the pre-processed LiDAR keeps the vertical information ($A_{LID_z}$ to $I_{LID_k}$) and both RGB and DVS images are naturally at the LiDAR's z-axis, the performance on DE and SS are getting improved. Although there is no specific transformation applied to the network architecture, the 15L model can learn the relationship between shared feature maps. This means that the LiDAR also plays an important role in DE and SS tasks and shows that the 15L model successfully leverages shared feature maps through intermediate fusion at the $1^{st}$ bottleneck.

Furthermore, based on the qualitative results shown in Figure \ref{fig:infer1} (rainy night) and Figure \ref{fig:infer2} (sunny day), the image quality of both 1L and 15L models are comparable on both DE and LS tasks. To be more specific on a sunny day (samples from nuScenes-lidarseg), the 15L model performance is quite similar to the 1L model. However, if we take a close look at the rear SS image, the 15L model can segment temporary road barriers successfully while the 1L model cannot. Besides that, on a rainy night (samples from set C), the 15L model is performing better where it can segment the road lane on the front view SS image. It is also better in recognizing surrounding vehicles on both BEVP images. 


\subsection{Static vs Adaptive Loss Weighting}

\begin{figure*}
	\centering

	\includegraphics[width=\textwidth]{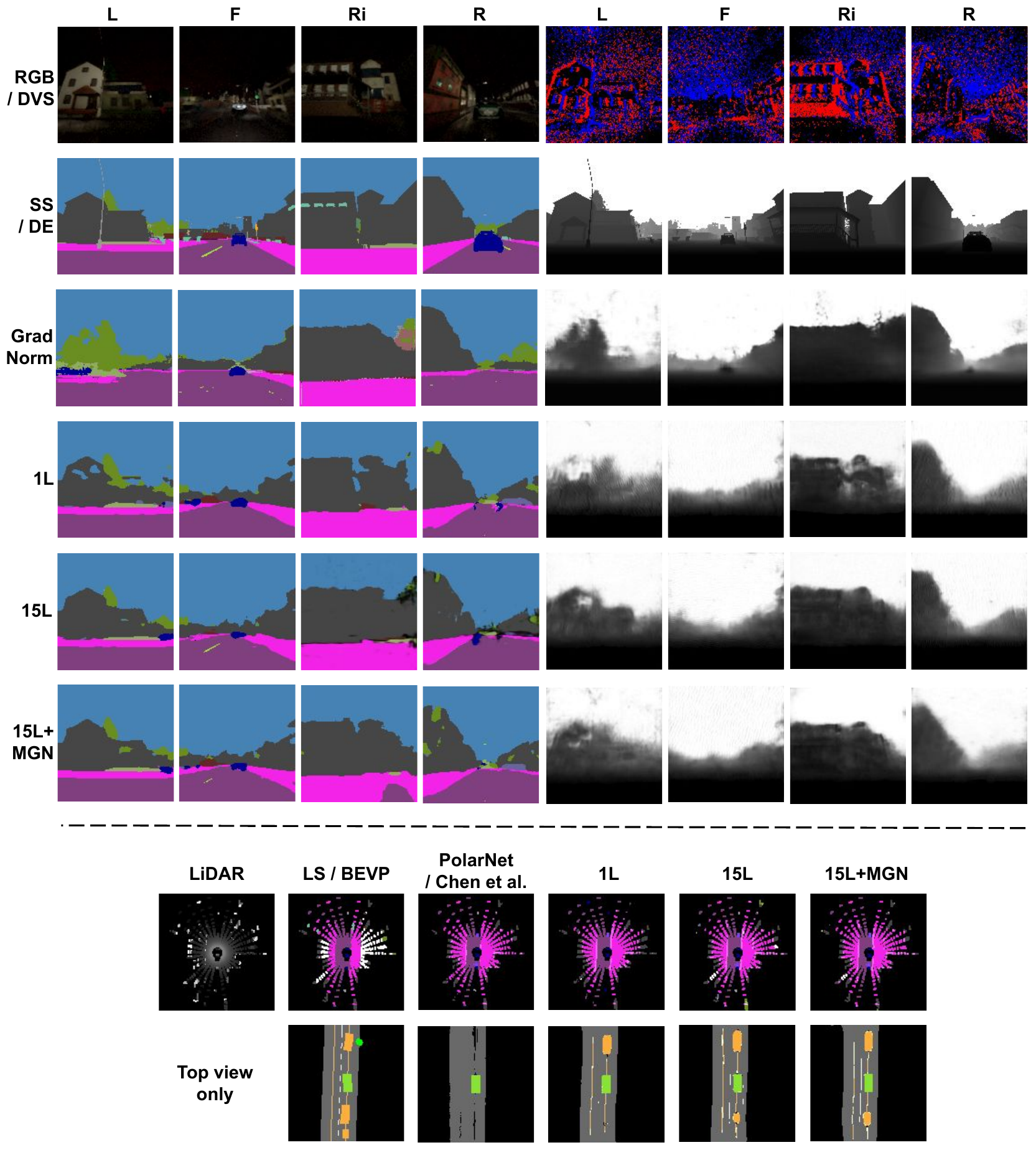}  

	\caption{Inference results on a rainy night (Test set C). A qualitative comparison between our model variants (1L, 15L, and 15L+MGN) and combination of STL (single-task learning) and MTL models by Chen et al.'s\cite{birdview3} (BEVP), PolarNet\cite{polarnet} (LS), and GradNorm\cite{adaptloss:gradnorm} (DE and SS).}
	\label{fig:infer1}
\end{figure*}

\begin{figure*}
	\centering

	\includegraphics[width=\textwidth]{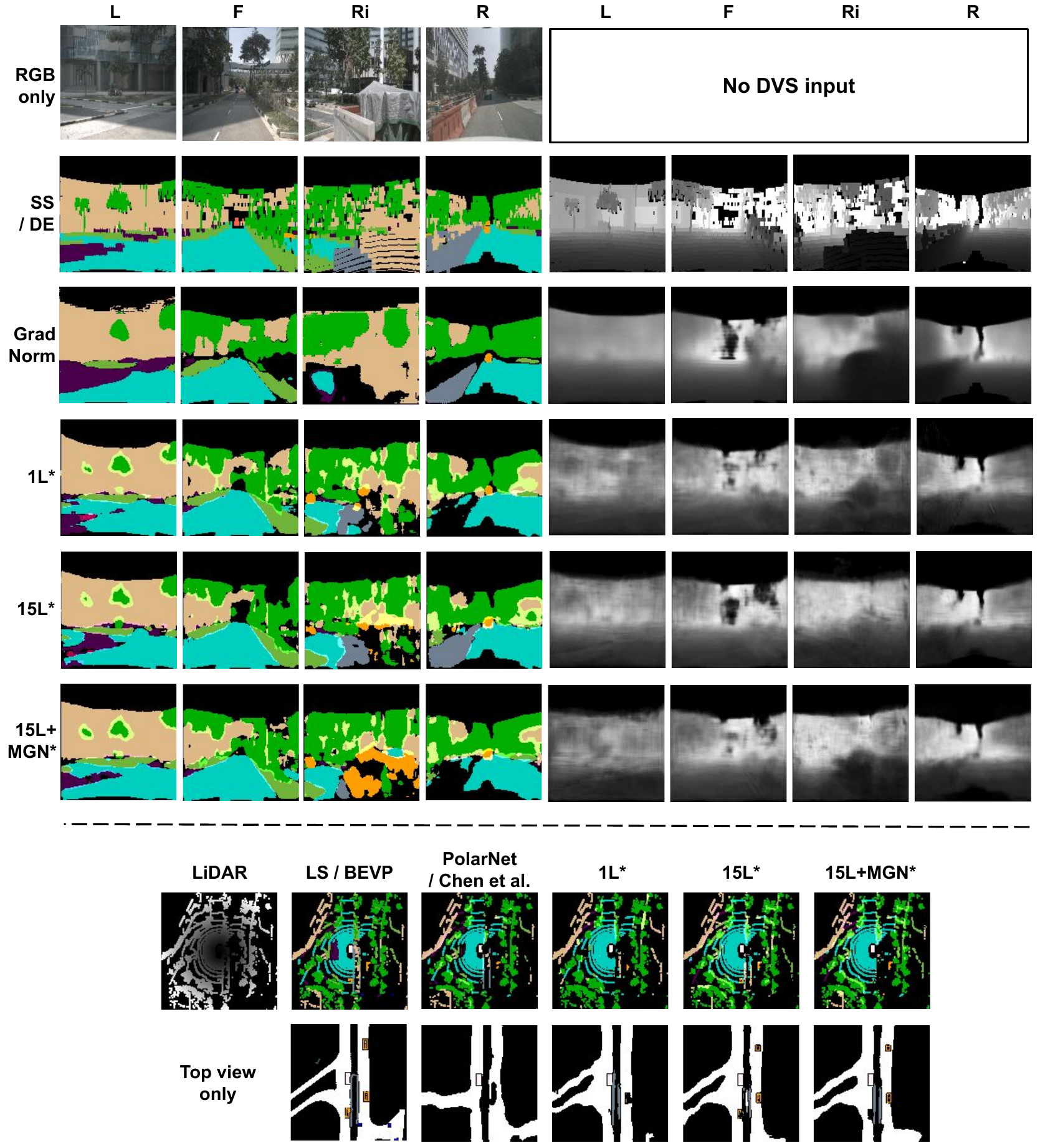}  

	\caption{Inference results on a sunny day (Test set nuScenes-lidarseg). A qualitative comparison between our model variants without DVS inputs (1L*, 15L*, and 15L+MGN*), and combination of STL (single-task learning) and MTL models by Chen et al.'s\cite{birdview3} (BEVP), PolarNet\cite{polarnet} (LS), and GradNorm\cite{adaptloss:gradnorm} (DE and SS).}
	\label{fig:infer2}
\end{figure*}

Plenty of tasks can lead to an uneven loss value which depends on what kind of loss function is used. Even if multiple related tasks are handled with the same loss function, it still can lead to imbalanced learning due to the different number of elements and characteristics at the output layer. For instance, the elements in $I_{LS}$ are much larger than in $I_{BEVP}$ and significantly different from $I_{SS}$. Therefore, a proper set of loss weights is needed to balance the task learning process. Moreover, it has to be tuned automatically to avoid an expensive computational cost in finding the best combination. Therefore, we propose the MGN algorithm to balance the rate of task learning by tuning each task’s loss weight adaptively.

Based on Table \ref{tab:testcompare}, the model trained with the modified GradNorm (MGN) algorithm (15L+MGN) has a better performance compared to the previous best model with static loss weights (15L) on all test sets. With a consistent result, the total metric (TM) score gets lowered from 1.448 to 1.393 (set A), 1.131 to 1.086 (set B), 1.050 to 0.999 (set C), and 1.079 to 1.069 (nuScenes-lidarseg). However, even with lower TM scores, not all tasks are getting improved by the model. The 15L+MGN variant may have better performance on LS and BEVP tasks where $IoU_{LS}$ and $IoU_{BEVP}$ scores are higher than the 15L variant. However, the 15L model still performs better than the 15L+MGN by achieving lower $MAE_{DE}$ and higher $IoU_{SS}$ on DE and SS tasks respectively. To be noted, the goal of the MGN algorithm is to improve the overall model performance by balancing the rate of learning on each task. Instead of improving the performance of each task, the MGN algorithm is focused on balancing the gradient signal among the tasks. Therefore, the 15L model may still have a better performance on some tasks. In this case, there is a performance trade-off, especially between DE-SS tasks and LS-BEVP tasks. Besides the TM score (\ref{eq:totmetric}), the metric variance (MV) (\ref{eq:metricvar}) can be used to determine the model performance based on the rate of discrepancy between tasks. As can be seen in Table \ref{tab:testcompare}, the MV of the 15L+MGN model is smaller than the 15L model on all simulation datasets and has the same MV on nuScenes-lidarseg. A lower MV indicates that the model performance on overall tasks is getting balanced with just a little discrepancy.

Based on the qualitative results shown in Figure \ref{fig:infer1} (rainy night) and Figure \ref{fig:infer2} (sunny day), we can see that the 15L+MGN model has a better BEVP performance where it has a more clear projection of a car behind the ego vehicle (sample of a rainy night in dataset C) and a better projection of the roadmap (sample of a sunny day in nuScenes-lidarseg). Meanwhile, the 15L model has a better SS performance on overall views. To be more specific, the 15L model can segment the sidewalk in the right SS image (set C) and the temporary road barriers (nuScenes-lidarseg).

\subsection{Loss Weighting Behavior}

\begin{figure}
	
			\begin{subfigure}{\linewidth}
				\centering
				\stackunder[5pt]{\centering\includegraphics[width=0.745\linewidth]{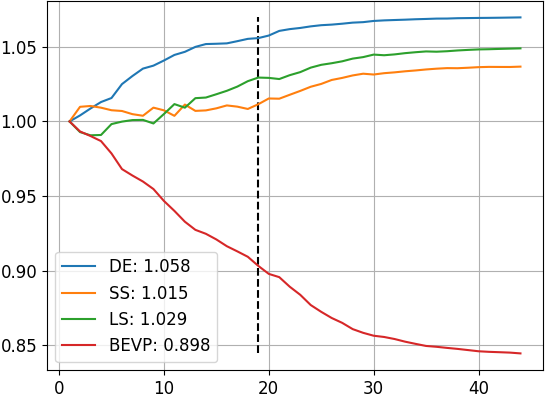}}{Loss weights update on dataset A}
			\end{subfigure}
			\par\bigskip
			
			\begin{subfigure}{\linewidth}
				\centering
				\stackunder[5pt]{\centering\includegraphics[width=0.745\linewidth]{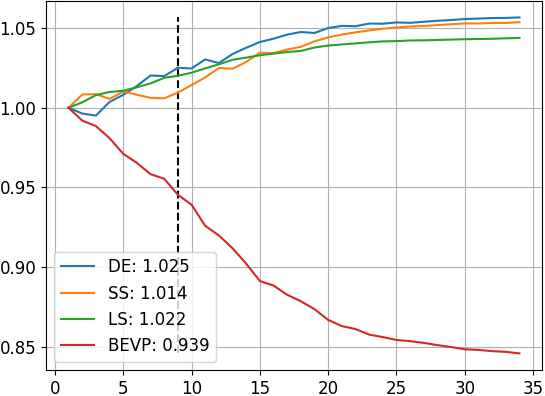}}{Loss weights update on dataset B}
			\end{subfigure}
			\par\bigskip
			\begin{subfigure}{\linewidth}
				\centering
				\stackunder[5pt]
				{\centering\includegraphics[width=0.745\linewidth]{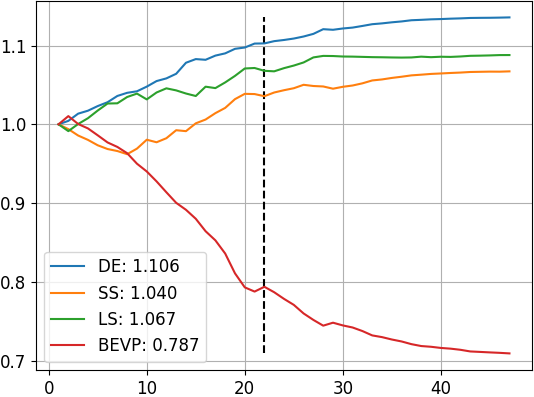}}{Loss weights update on dataset C}
			\end{subfigure}
			\par\bigskip
			
			\begin{subfigure}{\linewidth}
				\centering
				\stackunder[5pt]
				{\centering\includegraphics[width=0.745\linewidth]{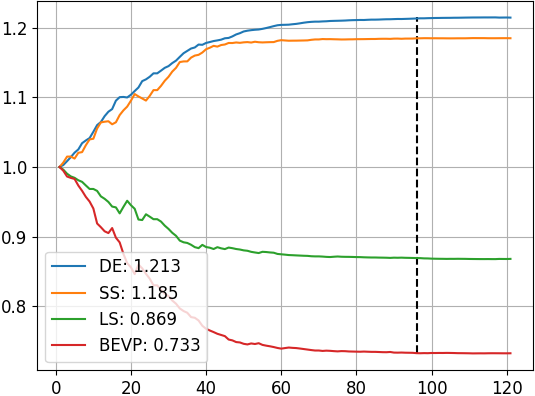}}{Loss weights update on nuScenes-lidarseg}
			\end{subfigure}
		\caption{Loss weights update. The vertical black dashed line shows the exact epoch where the model is converged with the loss weights printed on the legend. The vertical axis on each image is the loss weight while the horizontal axis is the epoch.}
		\label{fig:adaptlwprocess}
	\end{figure}

The loss weights update process of the 15L+MGN model during the training phase can be seen in Figure \ref{fig:adaptlwprocess}. In all simulation datasets, at the time when the model is converged, the modified GradNorm (MGN) algorithm tends to have similar behavior where it gives the highest loss weight to the depth estimation (DE) task followed by LiDAR segmentation (LS) at the second, semantic segmentation (SS) at the third, and bird's eye view projection (BEVP) at the last. Meanwhile, the order between LS and SS is swapped in nuScenes-lidarseg. From the loss weights change behavior, the MGN is penalizing less on the BEVP task so that it will not cause a high imbalance. As shown in the network architecture on Figure \ref{fig:netarch}, the BEVP decoder is placed after the $2^{nd}$ bottleneck meaning that the network has more focus during training. The gradient produced from BEVP also affects the decoder of SS, DE, and LS tasks on the previous layer. Thus, it makes sense that the MGN penalizes the BEVP task less than the other tasks. For tasks that are placed before the $2^{nd}$ bottleneck, the DE task loss calculated with Huber loss function (\ref{eq:huberloss1}) produce a smaller loss value compared to LS and SS losses which are calculated by BCEDice loss function (\ref{eq:bcedice}). Hence, the MGN algorithm is likely giving higher loss weight to the DE task to compensate for the imbalance. With this mechanism, the network will not lose its focus on the DE task learning while still maintaining progress on learning the other tasks.

Moreover, even if the number of elements in $I_{LS}$ is equal to $I_{SS}$, both tasks have a significant difference in characteristics. In $I_{LS}$, many elements are filled with 0 since there are plenty of vacant points which are not getting scanned by LiDAR lasers. Meanwhile, in $I_{SS}$, many elements are filled with 1 representing the one-hot object class on each tensor channel and have a strong correlation as all points are captured by the RGB camera. As a result, $\mathcal{L}_{SS}$ tends to be bigger than $\mathcal{L}_{LS}$ even when computed with the same loss function. Therefore, to balance the task learning, MGN is giving more weight on the LS task than on the SS task. However, the loss weighting behavior on SS and LS tasks is not consistent in the real-world dataset. The weight order is swapped between those tasks as the characteristic nuScenes-lidarseg is different from the simulation dataset. To be noted, the sum of all loss weights will always equal to $T=4$ as they are normalized with (\ref{eq:lwnormalization}) at the end of each epoch.

\subsection{Comparison with Recent Models}


We also conduct further model testing by comparing our models with some recent models for each task. In the multi-task depth estimation (DE) and semantic segmentation (SS) comparison, we train the original GradNorm model\cite{adaptloss:gradnorm} for each view so that there are four models in total. We use the GradNorm SegNet\cite{segnet} version with VGG16 encoder \cite{vgg16} and symmetric task decoders for comparative study as the GradNrom authors only use this model for in-depth analysis. The training setup is configured to be the same as described in the GradNorm paper where Adam optimizer\cite{optim_adam} along with pixel-wise cross-entropy and squared losses are used to train the model. Then, in the LiDAR segmentation (LS) comparison, we train PolarNet\cite{polarnet} to take our pre-processed LiDAR point clouds. We use the same training configuration written in the provided code as the author did not mention the detail in their paper. Concisely, Adam optimizer\cite{optim_adam} along with cross-entropy loss is used to train the model until convergence. Finally, in the bird's eye view projection (BEVP) comparison, we replicate Chen et al.'s model\cite{birdview3} that takes front RGB and top view LS images as the input. The model has a BEVP decoder to perform BEVP and input reconstruction modules that reconstruct the front RGB and LS images. However, LS input and BEVP output are represented in RGB image representation $\{0,...,255\}^{3 \times 128 \times 128}$. This kind of representation is not suitable for segmentation-related tasks. Thus, we change it into a one-hot encoded image, so that each of them is represented as $\{0,1\}^{C \times 128 \times 128}$ where $C$ is the number of possible classes. Then, we put 1 extra point-wise $(1\times1)$ convolution layer and a sigmoid activation at the last layer of the LS input reconstruction module and BEVP decoder. With this modification, the metric function IoU (\ref{eq:iou}) can be calculated for comparison purposes. Furthermore, besides comparing all metric scores, we also compare the number of model parameters, model size, GPU memory utilization, and inference speed to measure how efficient the model is.

Based on Table \ref{tab:testcompare}, our best model variant (15L+MGN) is better than the combination of Chen et al., PolarNet, and GradNorm models. In small and large datasets, the 15L+MGN model has lower total metric (TM) scores of 1.393 (set A), 1.086 (set B), and 1.069 (nuScenes-lidarseg). Meanwhile, the other variants still maintain a comparable performance with a small gap. However, in the medium dataset (set C), the combination has a better performance with a TM score of 0.976. Independently, PolarNet consistently gives a better LS performance by achieving the highest $IoU_{LS}$ in all datasets. Meanwhile, Chen et al. and GradNorm models are still comparable in BEVP, DE, and SS tasks. However, based on the qualitative results shown in Figure \ref{fig:infer1} (rainy night) and Figure \ref{fig:infer2} (sunny day), both Chen et al. and GradNorm models are missing the surrounding vehicles. On BEVP images, Chen et al.'s model is able to locate the occupied area by the surrounding vehicles, but cannot segment the vehicle correctly (set C). It also cannot project the local roadmap as well as our models (nuScenes-lidarseg). Then, as shown on SS images, GradNorm is facing difficulties in segmenting vehicles on a rainy night. On the other hand, our model faces the same difficulties, but it can locate the occupied region properly. During a rainy condition, the DVS sensor is distracted by the raindrops. As a result, the DE performance of our model is getting degraded. Then, as shown on LS images, both PolarNet and our models have a similar performance where both models can segment all objects from the top view perspective and locate the corresponding pixel class nearly the same as in the ground truth. Then, based on Table \ref{tab:compareinfer}, our models have much fewer parameters where they only have less than 2\% of the total parameters owned by the combination. Even with that small number of parameters, the 15L+MGN model maintains a better performance with less GPU memory utilization. Hence, it has a smaller size and can infer faster with a speed of around 54 frames per second (FPS) on simulation datasets and 65 FPS on a real-world dataset. Considering the performance result shown in Table \ref{tab:testcompare} and Table \ref{tab:compareinfer} along with the qualitative result shown on Figure \ref{fig:infer1} (rainy night) and Figure \ref{fig:infer2} (sunny day), it can be said that our model is better and more efficient than the combination. Moreover, our model is more preferable due to its compactness and smaller size.

\begin{table*}
	\begin{center}
		\begin{tabular}{*8c}
			\toprule
			Dataset & Model & Parameters & Total Parameters & Size & Total Size & GPU Usage & FPS\\
			\midrule
			& Chen et al.\cite{birdview3} & 10365211 &  & 83.061 &  &  & \\
			& PolarNet\cite{polarnet} & 13403701 & 136728752 & 107.342 & 1094.867 & 1987 & 44.472\\
			Simulation & 4$\times$GradNorm$^{\dagger}$\cite{adaptloss:gradnorm} & 4$\times$28239960 &  & 4$\times$226.116 &  &  & \\
			(A, B, C) & 1L & 2519488 & 2519488 & 20.549 & 20.549 & 1049 & 57.117\\
			& 15L & 2521504 & 2521504 & 20.565 & 20.565 & 1049 & 56.018\\
			& 15L+MGN & 2521504 & 2521504 & 20.565 & 20.565 & 1049 & 54.428\\
			\hdashline
			& Chen et al.\cite{birdview3} & 10372539 &  & 83.120 &  &  & \\
			& PolarNet\cite{polarnet} & 13404286 & 136757437 & 107.347 & 1095.099 & 2025 & 48.236\\
			nuScenes & 4$\times$GradNorm$^{\dagger}$\cite{adaptloss:gradnorm} & 4$\times$28245153 &  & 4$\times$226.158 &  &  & \\
			-lidarseg & 1L* & 2275604 & 2275604 & 18.557 & 18.557 & 1015 & 65.969\\
			& 15L* & 2277620 & 2277620 & 18.574 & 18.574 & 1017 & 65.426\\
			& 15L+MGN* & 2277620 & 2277620 & 18.574 & 18.574 & 1017 & 65.192\\
			\bottomrule 
		\end{tabular}
	\end{center}
	\caption{Model Comparison}
	\label{tab:compareinfer}
	\begin{tablenotes}
		\small\item
		$^{\dagger}$The number of parameters and model size of the GradNorm model are multiplied by 4 as there are 4 models in total.
		\item *The model does not have DVS encoders as there is no DVS input given.\\
		For a fair comparison, we use the same GPU device (NVIDIA GTX 1080 Ti) to run all models with batch size = 1. However, the inference speed measured in frame per second (FPS) is slightly different on each dataset due to the fluctuating GPU performance. Therefore, we average the FPS over A, B, and C datasets for the inference on simulation data. We separate the measurement on nuScenes-lidarseg as it has different characteristics. Thus, there is a small change in the number of parameters, model size (in MB), and GPU usage (in MB).
	\end{tablenotes}
\end{table*}

The reason why our model can outperform the combination even with fewer parameters is that it can leverage feature sharing on its encoders that process multiple views of input to efficiently learn the features. The network architecture makes it possible for each decoder to take the advantage of the extracted features from each encoder. Besides that, our proposed techniques are also playing an important key in boosting the model performance and keeping the performance balanced. Based on Table \ref{tab:testcompare}, without using these methods, our model cannot be better than the combination. Our 15L+MGN model may win on datasets A, B, and nuScenes-lidarseg but lose on dataset C with a TM score gap of 0.023. However, if we take a close look at the scores on dataset C, the lowest TM score obtained by the combination is mostly influenced only by the outstanding performance of PolarNet which achieves $IoU_{LS}$ of 0.735. In fact, PolarNet has always maintained to be the best on the LS task in all datasets. In dataset C, PolarNet outperforms our model with an $IoU_{LS}$ gap of 0.057 which is the largest amongst all experiments. If the gap on $IoU_{LS}$ is similar to the gap in other datasets, our model might have won on dataset C. PolarNet has what is called “ring-connected CNN“ that is specifically used to process LiDAR data. Therefore, with a larger number of learnable parameters, it is more capable of capturing more useful features in varying areas. To be noted, based on the number of towns used on each dataset, it can be said that dataset C is more varied as it contains 5 different maps while datasets A, B, and nuScenes-lidarseg only 2 maps, and both are similar to each other.

\section{Conclusion} \label{conclusion}
In this research, we develop a compact deep multi-task learning (MTL) model to perform various driving perception tasks simultaneously in one forward pass. Through data pre-processing and multi-sensor fusion techniques, the model can process and combine multiple input modalities. In addition, we propose an adaptive loss weighting algorithm to tackle the imbalanced learning issue and boost the overall performance. To understand the influence and behavior of our proposed methods, an ablation experiment is conducted by creating several variants. Finally, a comparative study against the combination of some recent models is conducted to clarify performance and efficiency.

From the ablation and comparative experiment results on both simulation and real-world datasets, we disclosed several findings as follows. First, we conclude that by keeping the height information of the LiDAR point clouds $A_{LID}$'s z-coordinate, the overall model performance is improved, especially in the LS task that has direct skip connections from the LiDAR encoder. This is proven by the 15L model that has a better performance compared to the 1L model. Moreover, with rich vertical features given from the LiDAR encoder through intermediate fusion at the first bottleneck, the 15L model gains better performance on DE and SS tasks. Second, by using the modified GradNorm (MGN) algorithm to update the loss weights adaptively based on the gradient signal during training, the MTL process is successfully balanced. From the comparison result, the 15L+MGN model performs better than the 15L model where it has lower total metric (TM) and metric variance (MV) scores. To be more detailed, the MGN algorithm makes a trade-off between DE-SS tasks with LS-BEVP tasks. Based on the loss weighting behavior, the MGN algorithm tends to penalize less on the task that has a higher focus by default such as the bird's eye view projection (BEVP) task that is placed at the end of the network. This algorithm is also capable of compensating small or large losses produced by different loss functions with varying output elements. As evidence, the depth estimation (DE) loss computed with the Huber loss function has a bigger loss weight than LiDAR segmentation (LS) and semantic segmentation (SS) loss computed with the BCEDice loss function. Finally, based on the comparison against the combination of some recent models, our best model variant (15L+MGN) maintains better performance even with much fewer parameters. Hence, it can inference faster and consume less GPU memory which is more preferable for further deployment. 

In future works, research on adaptive network branching can be conducted by letting the model find the best architecture to avoid designing the network manually. Furthermore, tackling all perception, planning, and control tasks simultaneously is also an interesting challenge in autonomous driving research. Therefore, fully end-to-end learning of an autonomous driving agent can be achieved.


%





\ifCLASSOPTIONcaptionsoff
  \newpage
\fi



\bibliographystyle{IEEEtran}
\bibliography{references}
%




%

\newpage

\begin{IEEEbiography}[{\includegraphics[width=1in,height=1.25in,clip,keepaspectratio]{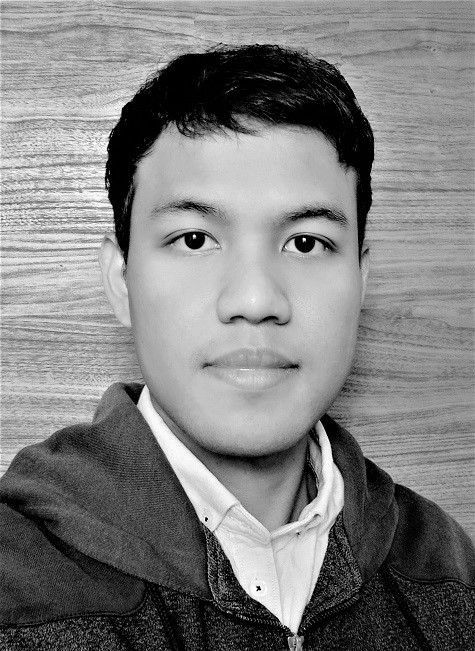}}]{Oskar Natan}
	received the B.A.Sc. degree in electronics engineering and M.Eng. degree in electrical engineering from Electronic Engineering Polytechnic Institute of Surabaya, Indonesia, in 2017 and 2019 respectively. Starting from January 2020, he has been a lecturer at the Department of Computer Science and Electronics, Universitas Gadjah Mada, Indonesia. Currently, Oskar is pursuing the Dr.Eng. degree at the Department of Computer Science and Engineering, Toyohashi University of Technology, Japan. 
\end{IEEEbiography}



\begin{IEEEbiography}[{\includegraphics[width=1in,height=1.25in,clip,keepaspectratio]{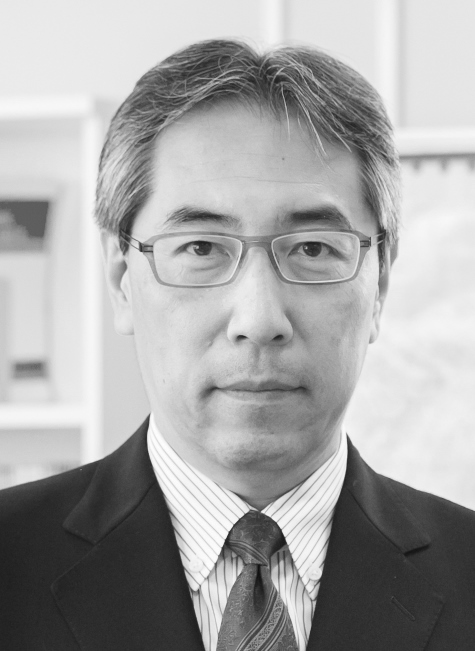}}]{Jun Miura}
	(Member, IEEE) received the B.Eng. degree in mechanical engineering, M.Eng. and Dr.Eng. degree in information engineering from The University of Tokyo, Japan, in 1984, 1986, and 1989 respectively. From 1989 to 2007, he was with the Department of Computer-controlled Mechanical Systems, Osaka University, Japan, first as a research associate and later as an Associate Professor. From March 1994 to February 1995, he was a visiting scientist at the Department of Computer Science, Carnegie Mellon University, USA. In 2007, he became a Professor at the Department of Computer Science and Engineering, Toyohashi University of Technology, Japan. To date, Professor Miura has published over 240 articles in the field of robotics and artificial intelligence in internationally reputable journals and conferences. He has received plenty of awards, including the Best Paper Award from the Robotics Society of Japan in 1997, the Best Paper Award Finalist at ICRA 1995, and the Best Service Robotics Paper Award Finalist at ICRA 2013. 
\end{IEEEbiography}



\vfill


\end{document}